\documentclass{article}

\PassOptionsToPackage{numbers}{natbib}

\usepackage[preprint]{neurips_2020}

\usepackage[T1]{fontenc}
\usepackage{hyperref}
\usepackage{url}
\usepackage{booktabs} 
\usepackage{amsfonts}
\usepackage{nicefrac}
\usepackage{microtype}
\usepackage{bm}
\usepackage{amsmath,amssymb}
\usepackage{amsthm}
\usepackage{caption}
\usepackage{graphicx}
\usepackage{subcaption}
\usepackage{algorithm,algorithmic}

\allowdisplaybreaks

\title{Bayesian Experience Reuse for Learning from Multiple Demonstrators}

\author{
  Michael Gimelfarb\thanks{Affiliate to Vector Institute of Artificial Intelligence, Toronto}\\
  Mechanical and Industrial Engineering \\ University of Toronto\\
  \texttt{mike.gimelfarb@mail.utoronto.ca}\\
  \And
   Scott Sanner \\
   Mechanical and Industrial Engineering \\ University of Toronto\\
   \texttt{ssanner@mie.utoronto.ca}\\
  \And
   Chi-Guhn Lee \\
   Mechanical and Industrial Engineering \\ University of Toronto\\  
   \texttt{cglee@mie.utoronto.ca}
}

\begin{document}

\maketitle

\begin{abstract}
    Learning from demonstrations (LfD) improves the exploration efficiency of a learning agent by incorporating demonstrations from experts. However, demonstration data can often come from multiple experts with conflicting goals, making it difficult to incorporate safely and effectively in online settings. We address this problem in the static and dynamic optimization settings by modelling the uncertainty in source and target task functions using normal-inverse-gamma priors, whose corresponding posteriors are, respectively, learned from demonstrations and target data using Bayesian neural networks with shared features. We use this learned belief to derive a quadratic programming problem whose solution yields a probability distribution over the expert models. Finally, we propose Bayesian Experience Reuse (BERS) to sample demonstrations in accordance with this distribution and reuse them directly in new tasks. We demonstrate the effectiveness of this approach for static optimization of smooth functions, and transfer learning in a high-dimensional supply chain problem with cost uncertainty.
\end{abstract}

\section{Introduction}

Learning from demonstrations (LfD) is a powerful approach to incorporate advice from experts in the form of demonstrations in order to speed up learning on new tasks. However, existing work in LfD assumes that demonstrations are generated from a single near-optimal agent with a single goal \citep{argall2009survey, chernova2014robot,schaal1997learning}. In practice, however, data may be available from multiple sub-optimal agents with conflicting goals. For example, when learning to operate a vehicle autonomously using human driver demonstrations \citep{bojarski2016end}, different drivers can have different goals and needs (e.g. destinations, priorities for safety or time) and experience levels. Relying on demonstrators whose goals are misaligned with the new target task can lead to unintended behaviors, which can be dangerous in performance-critical applications, and can be minimized by learning to trust the most relevant demonstrators.

In this paper, we focus on using LfD with multiple conflicting demonstrators for solving static and dynamic optimization problems. In order to measure the similarity between source and target task goals, we adopt a Bayesian framework. We first parameterize source and target task reward functions as linear functions in a common feature space, whose uncertainty is modeled using Normal-Inverse-Gamma priors that are updated to posterior distributions using source and target demonstrations. Taking a fully Bayesian treatment allows us to avoid premature convergence and is more robust to non-stationary data \citep{bishop2006pattern}. To jointly learn the posteriors and feature space, we build a Bayesian neural network with a shared encoder and multiple heads, one for each source and target task. We then formulate a quadratic programming problem (QP) whose solution yields a probability distribution over the source demonstrators. This allows us to trade-off the mean and variance of the uncertainty in source and target task goals in a principled manner.

In order to transfer demonstrations to new tasks in LfD, one approach is to pre-train the learner directly from the source data \citep{cruz2017pre}, or learn and reuse auxiliary representations from the source data such as policies \citep{fernandez2010probabilistic,taylor2011integrating}. However, the former can be ineffective when demonstrations assume conflicting goals, while meaningful policies may be difficult to solicit from the latter when demonstrators are limited, sub-optimal or exploratory in nature \citep{nicolescu2003natural,schaal1997learning,van2001learning}. On the other hand, experience collected from a failed or inexperienced demonstrator can be just as valuable as an experienced one \citep{grollman2011donut}. We present an algorithm called \emph{Bayesian Experience Reuse} (BERS), for directly reusing multiple demonstrations in a way that is consistent with the learned weighting over the source and target task goals (Algorithm~\ref{alg:main}). We demonstrate the efficacy of our approach for optimizing static functions, and continuous control of a supply chain using reinforcement learning with high-dimensional actions. 

\subsection{Related Work}

Most papers in LfD incorporate demonstrations from a single optimal or near-optimal expert with a single goal \citep{argall2009survey, chernova2014robot,schaal1997learning}. Some papers in the area of LfD for reinforcement learning (RLfD) relax this assumption to a single sub-optimal demonstrator, using pre-training \citep{cruz2017pre, gao2018reinforcement,hester2018deep}, reward shaping \citep{brys2015reinforcement,suay2016learning}, ranking demonstrations \citep{wang2017improving}, Bayesian inference \citep{price2003bayesian}, or other approaches \citep{taylor2011integrating} for transfer. However, the aforementioned papers cannot learn from multiple demonstrators with conflicting goals. On the other hand, papers on this topic typically assume multiple near-optimal demonstrators, so that recovering policies \citep{fernandez2010probabilistic,li2018reinforcement,li2018optimal}, value functions \citep{gimelfarb2018reinforcement,li2018oil} or successor features \citep{barreto2017successor,barreto2018transfer} is possible. In this paper, we show how \emph{sub-optimal} demonstrators with \emph{conflicting} goals can be ranked according to their alignment with the target task goal in a safe Bayesian manner, and reused directly by the target agent without learning auxiliary representations such as policies or value functions. Our approach is related to encoder sharing \citep{flet2019merl,perrone2018scalable,van2017hybrid,vezzani2019learning} and uncertainty quantification of rewards or policies \citep{azizsoltani2019unobserved, brown2020safe,janz2019successor,thakur2019uncertainty}, but our problem setting and methodology are different.

\section{Background}

\subsection{Reinforcement Learning}

Complex decision-making can be summarized in a \emph{Markov decision process} (MDP), formally defined as a five-tuple $\langle \mathcal{S}, \mathcal{A}, P, R, \gamma \rangle$, where: $\mathcal{S}$ is a set of states, $\mathcal{A}(s)$ is a set of possible actions in state $s$, $P(\cdot | s,a)$ gives the next-state distribution upon taking action $a$ in state $s$, $R(s,a,s')$ is the reward received upon transition to state $s'$ after taking action $a$ in state $s$, and $\gamma \in [0, 1]$ is a discount factor that gives smaller priority to rewards realized further into the future. The objective of an agent is to find a policy $\mu : \mathcal{S} \to \mathcal{A}$ that maximizes the long-run expected return $Q^\mu(s,a) = \mathbb{E}[\sum_{t=0}^\infty \gamma^t R(s_t,a_t,s_{t+1}) \,|\, s_0 =s, a_0 = a]$, where $a_t \sim \mu(s_t)$ and $s_{t+1} \sim P(\cdot | s_t, a_t)$. 

When $P$ and $R$ are known, an optimal policy can be found exactly using \emph{dynamic programming} \citep{powell2007approximate, puterman2014markov}. In the \emph{reinforcement learning} setting, neither $P$ nor $R$ are known to the decision-maker. Instead, the agent interacts with the environment using a randomized exploration policy $\mu^e$, collecting reward and observing the state transitions \citep{sutton2018reinforcement}. In order to learn optimal policies in this setting, \emph{temporal difference} methods first learn $Q(s,a)$ and then recover an optimal policy \citep{mnih2015human,watkins1992q}, while \emph{policy gradient} methods parameterize and recover an optimal policy $\mu^*$ directly \citep{sutton2000policy}. \emph{Actor-critic} methods learn and use both a critic $Q(s,a)$ and actor policy $\mu(s)$ simultaneously \citep{konda2000actor}. 

\subsection{Common Feature Representations} 

In this paper, we consider the regression problem associated with learning a multivariate function $y : \mathcal{X} \to \mathbb{R}$ on a domain $\mathcal{X}$. For instance, in the RL setting, we estimate reward functions where $\mathcal{X} = \mathcal{S} \times \mathcal{A} \times \mathcal{S}$ and $y(\mathbf{x}) = R(s, a, s')$. Given a feature map $\bm{\phi} : \mathcal{X} \to \mathbb{R}^d$, a function $y$ can be expressed as a linear combination $y(\mathbf{x}) = \bm{\phi}(\mathbf{x})^\top \mathbf{w}, \, \forall x \in \mathcal{X}$, where $\mathbf{w} \in \mathbb{R}^d$ is a fixed vector. 

We are interested in the transfer learning problem between similar tasks in a domain $\mathcal{M}^{\bm{\phi}}$ on a common $\mathcal{X}$, defined formally as
\begin{equation}
\label{eqn:domain}
    \mathcal{M}^{\bm{\phi}} = \left\lbrace y \,:\, \exists \mathbf{w} \in \mathbb{R}^d \textit{ s.t. } y(\mathbf{x}) = \bm{\phi}(\mathbf{x})^\top \mathbf{w},\,\forall \mathbf{x} \in \mathcal{X} \right\rbrace.
\end{equation}
In the RL setting, $\mathcal{M}^{\bm{\phi}}$ could include all MDPs with shared dynamics and different rewards $R_i$. In (\ref{eqn:domain}), we have explicitly assumed that the (unknown) state features $\bm{\phi}$ are shared among tasks. This is not a restrictive assumption in practice, as given a set of tasks $T_1, T_2\dots T_n \in \mathcal{M}^{\bm{\phi}}$, we may trivially define $\phi_{k}(\mathbf{x}) = y_{T_k}(\mathbf{x}),\,\forall \mathbf{x} \in \mathcal{X}$ for each $k = 1, 2 \dots n$. The main challenge in this paper is to simultaneously learn a suitable common latent feature embedding $\bm{\phi}$ and posterior distributions for $y(\mathbf{x})$, and leverage this uncertainty for measuring task similarity in $\mathcal{M}^{\bm{\phi}}$.

\section{Bayesian Learning from Multiple Demonstrators}

The agent is presented with sets of demonstrations $\mathcal{D}_1, \mathcal{D}_2, \dots \mathcal{D}_N$ from source tasks $T_1, T_2, \dots T_N \in \mathcal{M}^{\bm{\phi}}$, consisting of labeled pairs $(\mathbf{x}_t, y_t)$, and would like to solve a new target task $T_{target} \in \mathcal{M}^{\bm{\phi}}$. In order to make optimal use of the source tasks during training, the agent should learn to favor demonstrators whose underlying reward representation is \emph{closest} to the target reward.

\subsection{Bayesian Linear Regression with Common Feature Representations}

In order to tractably learn these reward representations, we learn a shared feature space $\hat{\bm{\phi}} \approx \bm{\phi}$, together with Bayesian regressors $\mathbb{P}(\mathbf{w}_i | \mathcal{D}_i)$ for the source task and a Bayesian regressor $\mathbb{P}(\mathbf{w}_{target} | \mathcal{D})$ for the target task such that $y_i(\mathbf{x}) = \bm{\phi}(\mathbf{x})^\top \mathbf{w}_i$ and $y_{target}(\mathbf{x}) = \bm{\phi}(\mathbf{x})^\top \mathbf{w}_{target}$ approximately hold for all $i$ and all realizations $\mathbf{x} \in \mathcal{X},\, y \in \mathbb{R}$. As we will soon show, the sharing of the features $\bm{\phi}$ is crucial to allow meaningful comparisons between source weights $\mathbf{w}_i$ and target weights $\mathbf{w}$. 

In order to tractably learn the features $\bm{\phi}$ as well as the corresponding 
posterior distributions, we parameterize $\bm{\phi}(\mathbf{x}) \approx \bm{\phi}_{\bm{\theta}}(\mathbf{x})$ using a deep neural network (encoder) with weight parameters $\bm{\theta}$, and model $\mathbf{w}_1,\dots \mathbf{w}_N, \mathbf{w}_{target}$ using the \emph{normal-inverse-gamma} (NIG) conjugate prior:
\begin{equation}
\begin{aligned}
    y_{i}(\mathbf{x}) = \bm{\phi}_{\bm{\theta}}(\mathbf{x})^\top \mathbf{w}_i + \varepsilon_i, \\
    \varepsilon_i \sim \mathcal{N}(0, \sigma_i^2), \quad\quad
    \mathbf{w}_i \sim \mathcal{N}(\bm{\mu}_i, \sigma_i^2 \bm{\Lambda}_i^{-1}), \quad\quad
    \sigma_i^2 &\sim \mathrm{InvGamma}(\alpha_i, \beta_i).
\end{aligned}
\label{eqn:nig}
\end{equation}
The joint posterior $\mathbb{P}(\mathbf{w}_i, \sigma_i^2 | \mathcal{D}_i)$ can be shown to factor as
\begin{equation}
\label{eqn:posterior}
    \mathbb{P}(\mathbf{w}_i, \sigma_i^2 | \mathcal{D}_i) \propto \mathbb{P}(\mathbf{w}_i | \sigma_i^2, \mathcal{D}_i)\, \mathbb{P}(\sigma_i^2 | \mathcal{D}_i),
\end{equation}
where
$\mathbf{w}_i | \sigma_i^2, \mathcal{D}_i \sim \mathcal{N}(\bm{\mu}_i, \sigma_i^2 \bm{\Lambda}_i^{-1})$ and $\sigma_i^2 | \mathcal{D}_i \sim \mathrm{InvGamma}(\alpha_i,\beta_i)$, where:
\begin{equation}
\begin{aligned}
    \bm{\Lambda}_i
    = \bm{\Lambda}_i^0 + \bm{\Phi}_i^\top \bm{\Phi}_i, &\quad\quad 
    \bm{\mu}_i
    =  \bm{\Lambda}_i^{-1} \left(\bm{\Lambda}_i^0 \bm{\mu}_i^0 + \bm{\Phi}_i^\top \mathbf{y}_i \right), \\
    \alpha_i = \alpha_i^0 + \frac{1}{2} |\mathcal{D}_i|, &\quad\quad
    \beta_i
    = \beta_i^0 + \frac{1}{2}\left(\mathbf{y}_i^\top \mathbf{y}_i + (\bm{\mu}_i^0)^\top \bm{\Lambda}_i^0 \bm{\mu}_i^0 - \bm{\mu}_i^\top \bm{\Lambda}_i \bm{\mu}_i \right),
\end{aligned}
\label{eqn:nig_update}
\end{equation}
and where $\bm{\Phi}_i$ is the set of state features and $\mathbf{y}_i$ is the set of observations of $y_i$ in $\mathcal{D}_i$ \citep{bishop2006pattern}. We have also assumed that, conditional on data $\mathcal{D}_i$, weights $\mathbf{w}_i$ and variances $\sigma_i^2$ are mutually independent between tasks, a very mild assumption in practice. Adapting the methodology of \citet{snoek2015scalable}, we update $\bm{\theta}$ by gradient ascent on the \emph{marginal log-likelihood} function for each head $i$, defined here as
\begin{equation}
\label{eqn:nig_ll}
    \log \mathbb{P}(\mathbf{y}_i | \mathbf{X}_i) = |\mathcal{D}_i|\pi +\log \Gamma(\alpha_i^0) - a_i^0 \log \beta_i^0 + \frac{1}{2}\log | \bm{\Lambda}_i^0| - \log\Gamma(\alpha_i) + \alpha_i \log\beta_i - \frac{1}{2} \log |\bm{\Lambda}_i|,
\end{equation}
where the key quantities are provided by (\ref{eqn:nig_update}) and depend implicitly on $\bm{\theta}$ through $\bm{\Phi}_i$. 

Using this model, parameter sharing allows the learned features $\bm{\phi}$ to be refined and transferred seamlessly from source to target tasks, or as new source and target tasks are incorporated, and provides an additional source of knowledge for transfer. However, our main contribution is to use the posterior distributions over task goals, $\mathbf{w}_i$ and $\mathbf{w}_{target}$, to transfer experiences between tasks. 

\subsection{Source Task Selection via Quadratic Programming}

In order to derive a Bayesian decision rule for source task selection, we first observe that source tasks that are most similar to the target task, and hence those that lead to better transfer, should have $\mathbf{w}_i$ closest to the true target $\mathbf{w}_{target}$. In our setting, we instead have uncertain estimates for $\mathbf{w}_i$ and $\mathbf{w}_{target}$ modelled as multivariate normal random variables. We therefore look for a weighting $\sum_{i=1}^N a_i \mathbf{w}_i$ that is \emph{closest in expectation} to $\mathbf{w}_{target}$ with respect to the uncertainty in these estimates.

More specifically, suppose that we have computed the posterior distributions of $\mathbf{w}_i$ for each $i = 1, 2 \dots N$ and $\mathbf{w}_{target}$ from past data. We would like to weight the $\mathbf{w}_i$ in such a way that the weighted sum $\sum_i a_i \mathbf{w}_i$ is closest to $\mathbf{w}_{target}$ in expectation, or in other words, solve
\begin{equation}
\label{eqn:objective}
    \min_{\mathbf{a} \in \mathcal{P}} \mathbb{E}\left[\|\mathbf{w}_{target} - \sum_{i=1}^N a_i \mathbf{w}_i \|_2^2 \,\Big|\, \mathcal{D}\right],
\end{equation}
where $\mathcal{D}$ is the union of all source and target data and $\mathcal{P}$ is a convex polyhedron. Specifically, we would like to be able to sample source tasks according to $\mathbf{a}$, so we take $\mathcal{P} = \lbrace \mathbf{a} : \mathbf{1}^\top \mathbf{a} = 1, \,\mathbf{a} \geq \mathbf{0} \rbrace$, the set of probability distributions on $\lbrace 1, 2 \dots N\rbrace$. In other applications, such as static regression problems \citep{pardoe2010boosting,wang2019transfer,yao2010boosting}, we may choose $\mathcal{P} = \mathbb{R}^N$ or put additional constraints on expert selection.

To derive a closed form for (\ref{eqn:objective}), first note that, given the true values of the variances $\mathbf{v} = [\sigma_1^2, \dots \sigma_N^2, \sigma_{target}^2]$, the random variables $\mathbf{w}_1, \dots \mathbf{w}_N, \mathbf{w}_{target}$ follow normal distributions with respective means $\bm{\mu}_1 \dots \bm{\mu}_N, \bm{\mu}_{target}$ and covariances $\sigma_1^2 \bm{\Sigma}_1, \dots \sigma_N^2 \bm{\Sigma}_N, \sigma_{target}^2 \bm{\Sigma}_{target}$ (equation (\ref{eqn:posterior})). Then, for any $\mathbf{a} \in \mathbb{R}^N$, it is easy to check that $\mathbf{w}_{target} - \sum_{i} a_i \mathbf{w}_i \,|\, \mathbf{v}, \mathcal{D} \sim \mathcal{N}(\bm{\mu}_{target} - \sum_i a_i \bm{\mu}_i, \sigma_{target}^2 \bm{\Sigma}_{target} + \sum_i a_i^2 \sigma_i^2 \bm{\Sigma}_i)$, and hence that
\begin{align}
    \mathbb{E} \Big[\|\mathbf{w}_{target} - \sum_{i} a_i \mathbf{w}_i \|_2^2 \,|\, \mathbf{v}, \mathcal{D} \Big] &= \mathrm{tr}(\sigma_{target}^2 \bm{\Sigma}_{target} + \sum_i a_i^2 \sigma_i^2 \bm{\Sigma}_i) + \|  \bm{\mu}_{target} - \sum_i a_i \bm{\mu}_i\|_2^2 \nonumber \\
    &= \sigma_{target}^2 \mathrm{tr}(\bm{\Sigma}_{target}) + \sum_i a_i^2 \sigma_i^2 \mathrm{tr}(\bm{\Sigma}_i) + \|  \bm{\mu}_{target} - \sum_i a_i \bm{\mu}_i\|_2^2.
    \label{eqn:expanded_qp1}
\end{align}
Then, taking expectation of (\ref{eqn:expanded_qp1}) with respect to the variance terms $\mathbf{v}$,
\begin{align}
    \mathbb{E}\Big[\|\mathbf{w}_{target} - \sum_{i} a_i \mathbf{w}_i \|_2^2 \,\big|\, \mathcal{D}\Big]
    &= \mathbb{E}\left[\mathbb{E}\Big[\|\mathbf{w}_{target} - \sum_{i} a_i \mathbf{w}_i \|_2^2 \,\big|\, \mathbf{v}, \mathcal{D}\Big] \,\Big|\, \mathcal{D} \right] \nonumber \\
    &\propto \sum_i a_i^2 \, \mathbb{E}[\sigma_i^2 | \mathcal{D}] \, \mathrm{tr}(\bm{\Sigma}_i) + \|  \bm{\mu}_{target} - \sum_i a_i \bm{\mu}_i\|_2^2 \nonumber \\
    &= \sum_{i=1}^N a_i^2 \left(\frac{\beta_i}{\alpha_i - 1}\right) \mathrm{tr}(\bm{\Sigma}_i) + \|  \bm{\mu}_{target} - \sum_{i=1}^N a_i \bm{\mu}_i \|_2^2,
    \label{eqn:expanded_qp2}
\end{align}
where in the last step we used (\ref{eqn:nig}) and the expectation of $\mathrm{InvGamma}(\alpha_i, \beta_i)$, and ignored all terms independent of any decision variables $\mathbf{a}$. Hence, we have shown that \emph{optimizing the expected weighted mean-squared error (\ref{eqn:objective}) is equivalent to optimizing the weighted mean-squared error of the posterior means plus a penalty term equal to the product of the noise and posterior variances}.

To simplify (\ref{eqn:expanded_qp2}) further, we define $\mathbf{M} = [\bm{\mu}_1 \, \dots \bm{\mu}_N] \in \mathbb{R}^{d \times N}$ and $\mathbf{S} \in \mathbb{R}^{N \times N}$ the diagonal matrix with entries $\frac{\beta_i}{\alpha_i-1}\mathrm{tr}(\bm{\Sigma}_i), \,i=1, 2\dots N$. Rewriting (\ref{eqn:expanded_qp2}) in this new notation, we can obtain the following \emph{quadratic programming} formulation of our problem:
\begin{equation}
\label{eqn:qp}
\begin{alignedat}{2}
&\!\min_{\mathbf{a}}        &\qquad&  -\bm{\mu}^\top \mathbf{M} \mathbf{a} +  \frac{1}{2} \mathbf{a}^\top (\mathbf{M}^\top \mathbf{M} + \mathbf{S}) \mathbf{a} \\
&\text{subject to} &      & \mathbf{1}^\top \mathbf{a} = 1, \\
& & &  \mathbf{a} \succeq \mathbf{0}.
\end{alignedat}
\end{equation}
Here, $\mathbf{M}^\top \mathbf{M} + \mathbf{S}$ is positive definite, since it is the sum of the positive semi-definite $\mathbf{M}^\top \mathbf{M}$ and the positive definite $\mathbf{S}$. Hence, the above QP problem can be formulated and solved exactly using an off-the-shelf solver in polynomial time in $N$ \citep{boyd2004convex}, which does not grow with the feature dimension $d$ nor the number of demonstrations $|\mathcal{D}_i|$ and can be applied in online settings. Hence, \emph{the QP (\ref{eqn:qp}) remains tractable when the domain complexity is high or the number of demonstrations is large}. This is not 
necessarily the case if omitting the second order terms $\mathbf{S}$, since the QP may become rank-deficient and thus lack a unique solution. We can therefore interpret $\mathbf{S}$ as a \emph{data-dependent Bayesian regularizer for the QP solution.}

In the case where $N$ is considerably large, the solution to (\ref{eqn:qp}) can instead be approximated using neural networks \citep{amos2017optnet,bouzerdoum1993neural}. Alternatively, since the posterior updates are relatively smooth (as we demonstrate experimentally), the solution of (\ref{eqn:qp}) should not change significantly between consecutive iterations, in which case warm starts may be particularly effective \citep{ferreau2008online}.

\subsection{Bayesian Experience Reuse (BERS) for LfD}

When transferring demonstrations from a single source, a simple yet effective approach is to pre-train or seed the target learning agent with the demonstrations prior to target task training \citep{cruz2017pre}. When data originates from multiple demonstrators with differing goals, pre-training is no longer possible. Instead, we propose to train the target agent on the source data while concurrently learning the target task. In this manner, the source demonstrations provide an effective exploration bonus by allowing the agent to observe demonstrations from other tasks with similar goals. 

More specifically, in each iteration of the target learning phase, we sample a source task $T_i \in \mathcal{M}^{\bm{\phi}}$ according to the probability distribution $\mathbf{a}$ obtained from the quadratic programming problem (\ref{eqn:qp}), and train the target agent on experiences drawn from the corresponding source data $\mathcal{D}_i$. In order for the target agent to improve beyond the source task solution and generalize to the new task, the agent must eventually learn from target demonstrations rather than the source data. In order to do this, we apply the framework of Probabilistic Policy Reuse \citep{fernandez2010probabilistic,li2018optimal} in our setting. Specifically in each training episode $m$, the target agent is trained on source data with probability $p_m$, and otherwise uses target data with probability $1 - p_m$, where $p_m$ is annealed to zero over time. The choice of $p_m$ is typically problem dependent, but may also be annealed based on the QP solution.

\begin{figure}[!htb]
    \begin{minipage}{0.5\linewidth}
    \includegraphics[width=0.99\linewidth]{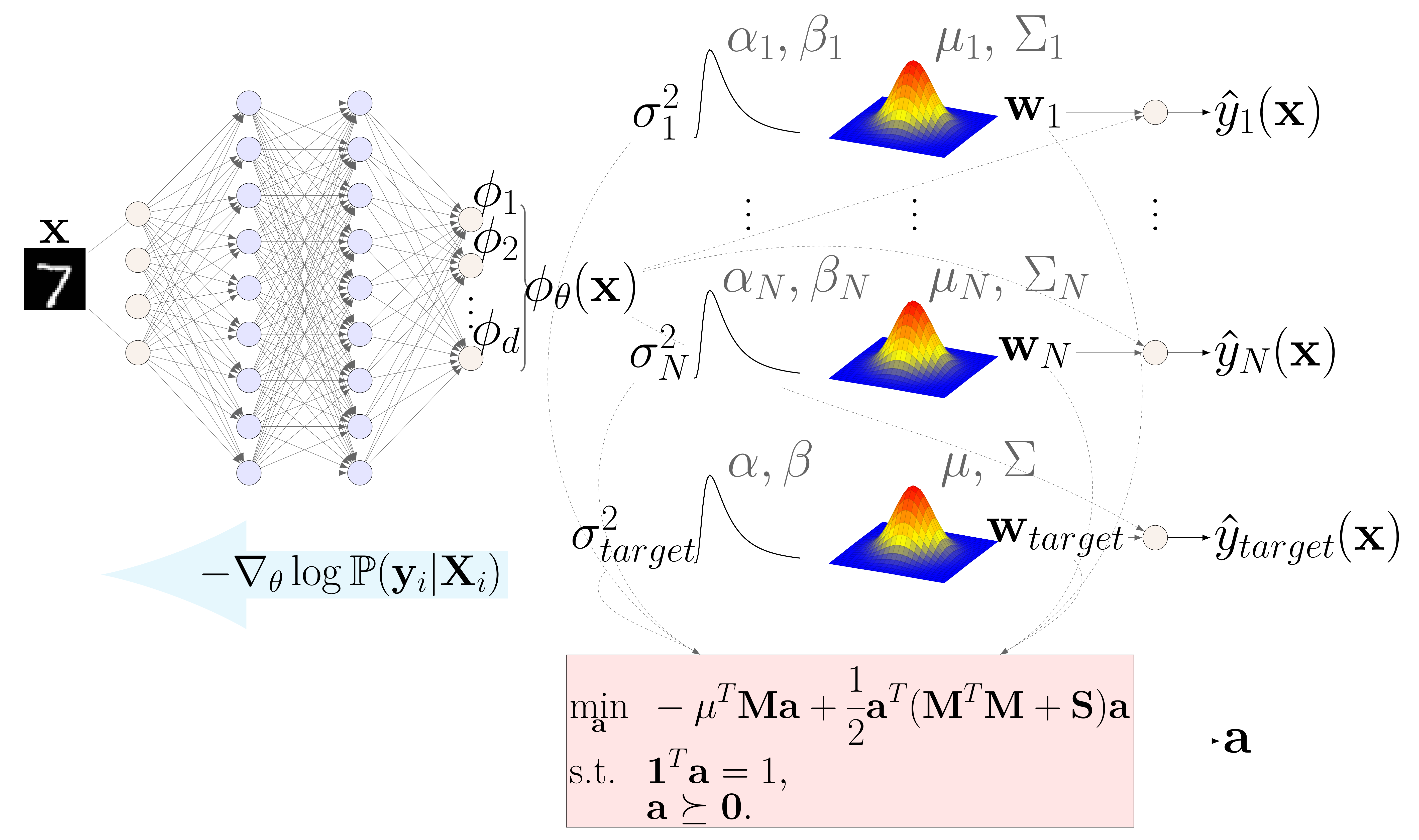}
    \label{fig:mhneural}
    \end{minipage} 
    \begin{minipage}{0.5\linewidth}
        \begin{algorithm}[H]
        \algsetup{linenosize=\tiny}
        \tiny
	    \caption{Bayesian Experience Reuse (BERS)}
    	\label{alg:main}
    	\begin{algorithmic}
    	    \REQUIRE $\lbrace \mathcal{D}_i \rbrace_{i=1}^N$, $T_{N+1} = T_{target} \in \mathcal{M}^{\bm{\phi}}$, $O_{base}$, $\mathcal{D}_{N+1} = \mathcal{D}_{target} = \emptyset$, $\bm{\theta}$, $\lbrace \bm{\mu}_i, \bm{\Lambda}_i, \alpha_1, \beta_i \rbrace_{i=1}^{N+1}$, $p_m$, $\mathbf{a}$
    	    \STATE pre-train $\bm{\theta}$, $\lbrace \bm{\mu}_i, \bm{\Lambda}_i, \alpha_1, \beta_i \rbrace_{i=1}^N$ on  $\lbrace \mathcal{D}_i \rbrace_{i=1}^N$ using (\ref{eqn:nig_update}), (\ref{eqn:nig_ll})
    		\FOR{episode $m=1, 2, \dots$}
    		\FOR{step $t=1, 2, \dots T$ of episode $m$}
    		    \STATE explore $T_{target}$ using $O_{base}$ and collect $d = (\mathbf{x}, y)$ 
    		    \STATE $\mathcal{D}_{target} = \mathcal{D}_{target} \cup d$ \IF{$\mathrm{Bernoulli}(p_m) = 1$}
    		        \STATE sample $i_t \sim \mathbf{a}$ and experience $\mathcal{B} \subset \mathcal{D}_{i_t}$
    		    \ELSE
    		        \STATE sample experience $\mathcal{B} \subset \mathcal{D}_{target}$
    		    \ENDIF
    		    \STATE train $O_{base}$ on $\mathcal{B}$
    		\ENDFOR
    		\STATE train $\bm{\theta}$, $\lbrace \bm{\mu}_i, \bm{\Lambda}_i, \alpha_1, \beta_i \rbrace_{i=1}^{N+1}$ on $\lbrace \mathcal{D}_i \rbrace_{i=1}^{N+1}$ using (\ref{eqn:nig_update}), (\ref{eqn:nig_ll})
    		\STATE solve QP (\ref{eqn:qp}) (using, e.g. \citep{andersen2015cvxopt}) and set $\mathbf{a}$ to the solution
    		\ENDFOR
    		\STATE \textbf{return} $O_{base}$
    	\end{algorithmic}
\end{algorithm}
     \end{minipage} 
    \caption{Left: bayesian multi-headed neural-linear model with shared encoder (MLP) and aggregated QP decision layer. Right: pseudo-code implementation for transfer learning for static and dynamic optimization (BERS).}
    \label{fig:code}
\end{figure}

The full implementation of our algorithm, \emph{Bayesian Experience Reuse} (BERS), is illustrated in Figure~\ref{fig:code} in a transfer learning setting. The left figure provides a visualization of the process used to weight source tasks. Here, Bayesian heads with parameters $\lbrace \bm{\mu}_i, \bm{\Lambda}_i, \alpha_i, \beta_i \rbrace_{i=1}^N$ and  $\bm{\mu}, \bm{\Lambda}, \alpha, \beta$ are maintained for source and target tasks respectively, while sharing the encoder parameters $\bm{\theta}$, and then used to construct the quadratic programming problem (\ref{eqn:qp}). The outputs $\hat{y}_i$ and $\hat{y}_{target}$ can also be used for making predictions (e.g. in regression tasks); in this paper we focus on using the posterior distributions over task goals, and the corresponding QP solution, to transfer source demonstrations in a Bayesian framework. 

In the pseudo-code on the right, we have defined $O_{base}$ to be a base learning algorithm for solving the target task, which is assumed to be a static or dynamic (RL) optimization algorithm in this work. We pre-train the first $N$ heads on source data to learn features and posteriors for source task goals. At the end of each episode, we train all source and target heads on source and target demonstrations to refine the features and posteriors for all tasks. With simple modifications, this algorithm can also be applied in multi-task settings, by maintaining one QP solution per task. The bottleneck of this procedure is the solution of the QP, which can be addressed using various approximation methods outlined above.

\section{Empirical Evaluation}

In order to demonstrate the effectiveness of BERS, we apply it to a static optimization problem and continuous control of a supply chain.

\subsection{Static Function Optimization}

We first consider the problem of minimizing a smooth function \citep{jamil2013literature}. Transfer learning can be useful in this setting because the known solution of one smooth function can be used as a starting point or seed in the search for the minimum of another similar function. More specifically, we use the 10-D Rosenbrock, Ackley and Sphere functions as source tasks, whose global optimum solutions have been shifted to various locations. As target task, we use each source task as the ground truth as well as the Rastrigin function. The Sphere and Rastrigin functions are locally similar around the optimum, so a successful transfer experiment should learn to transfer knowledge between them. We consider optimizing the Rastrigin function in both transfer and multi-task settings. As the base learning agent $O_{base}$ in Algorithm~\ref{alg:main}, we use Differential Evolution (DE) \citep{storn1997differential}, a commonly-used evolutionary algorithm for global optimization of black-box functions \footnote{Another option is to use Bayesian optimization (BO). However, we believe this is more suitable for expensive functions with relatively low numbers of local optima, such as for hyper-parameter optimization.}.

\begin{figure}[!htb]
    \centering
    \begin{subfigure}{.195\textwidth}
        \centering
        \includegraphics[width=0.99\textwidth]{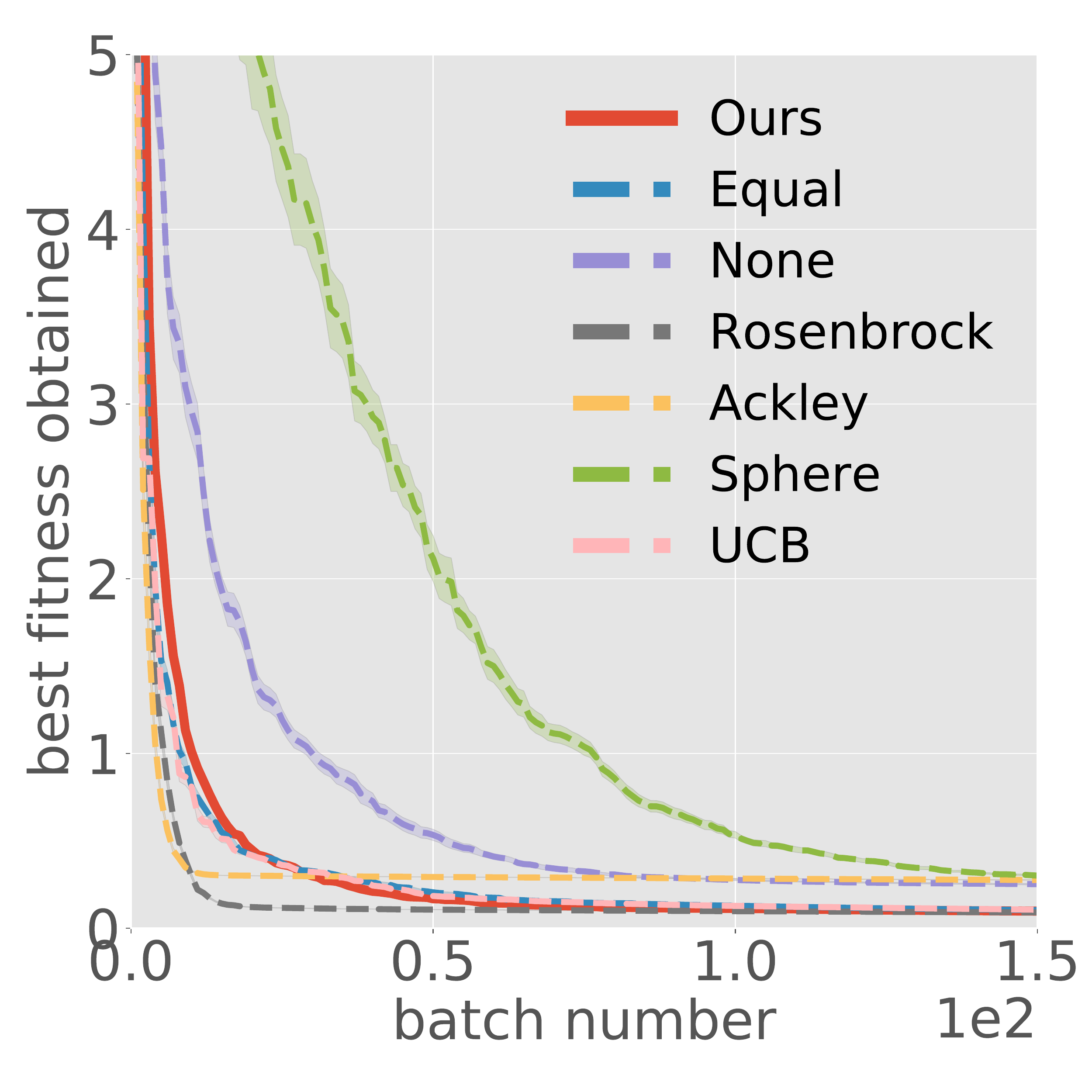}
        \includegraphics[width=0.99\textwidth]{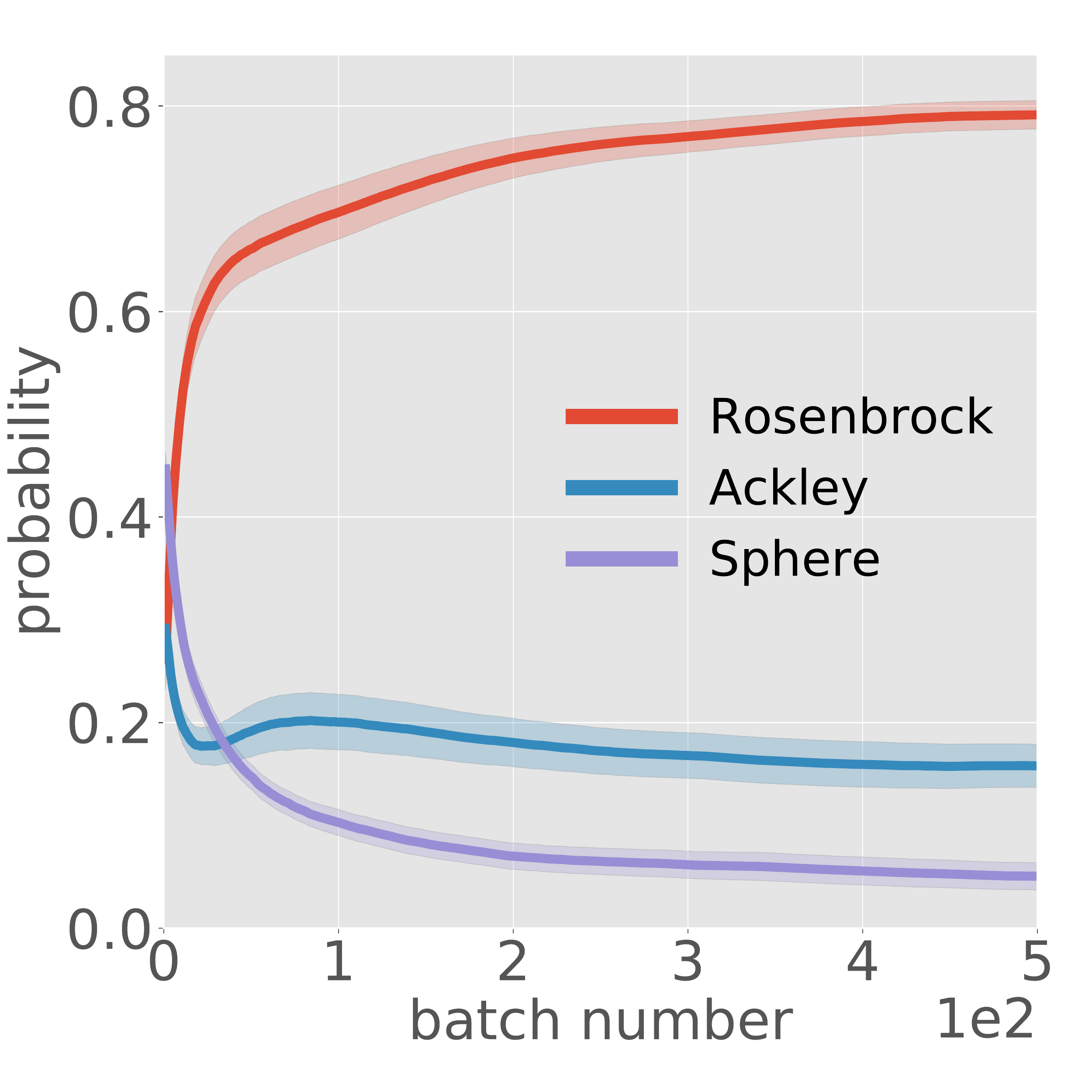}
        \caption{Rosenbrock}
    \end{subfigure}%
    \begin{subfigure}{.195\textwidth}
        \centering
        \includegraphics[width=0.99\textwidth]{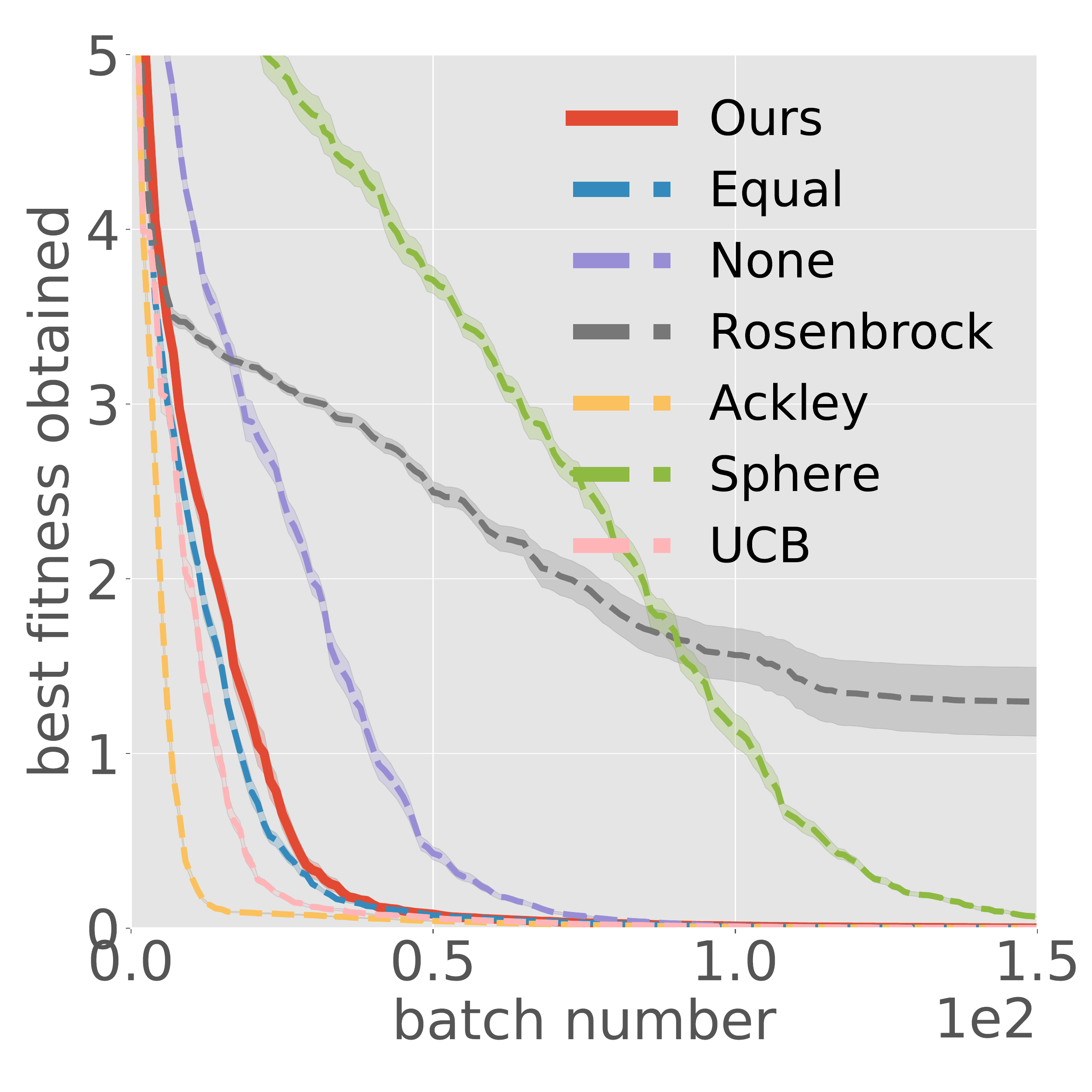}
        \includegraphics[width=0.99\textwidth]{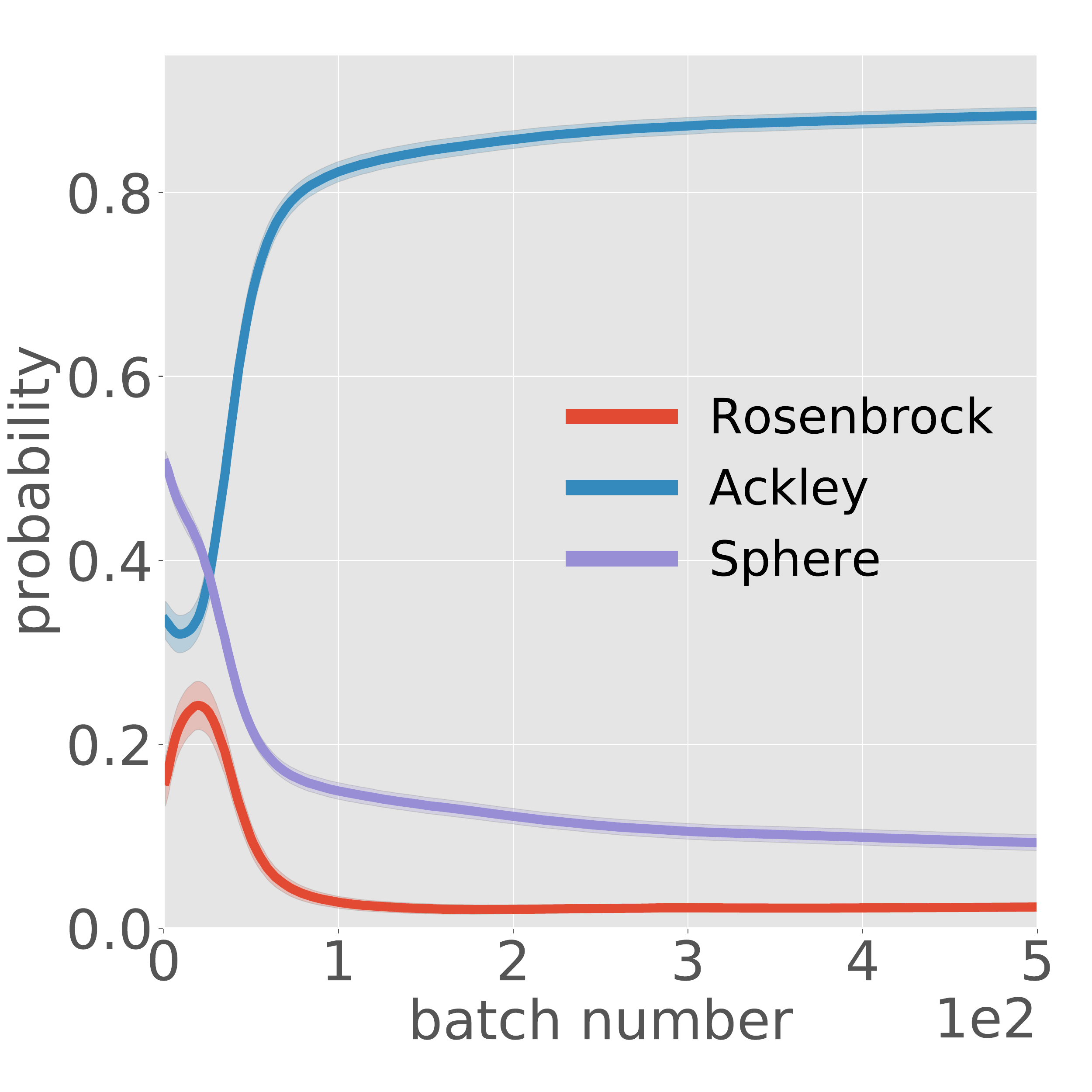}
        \caption{Ackley}
    \end{subfigure}%
    \begin{subfigure}{.195\textwidth}
        \centering
        \includegraphics[width=0.99\textwidth]{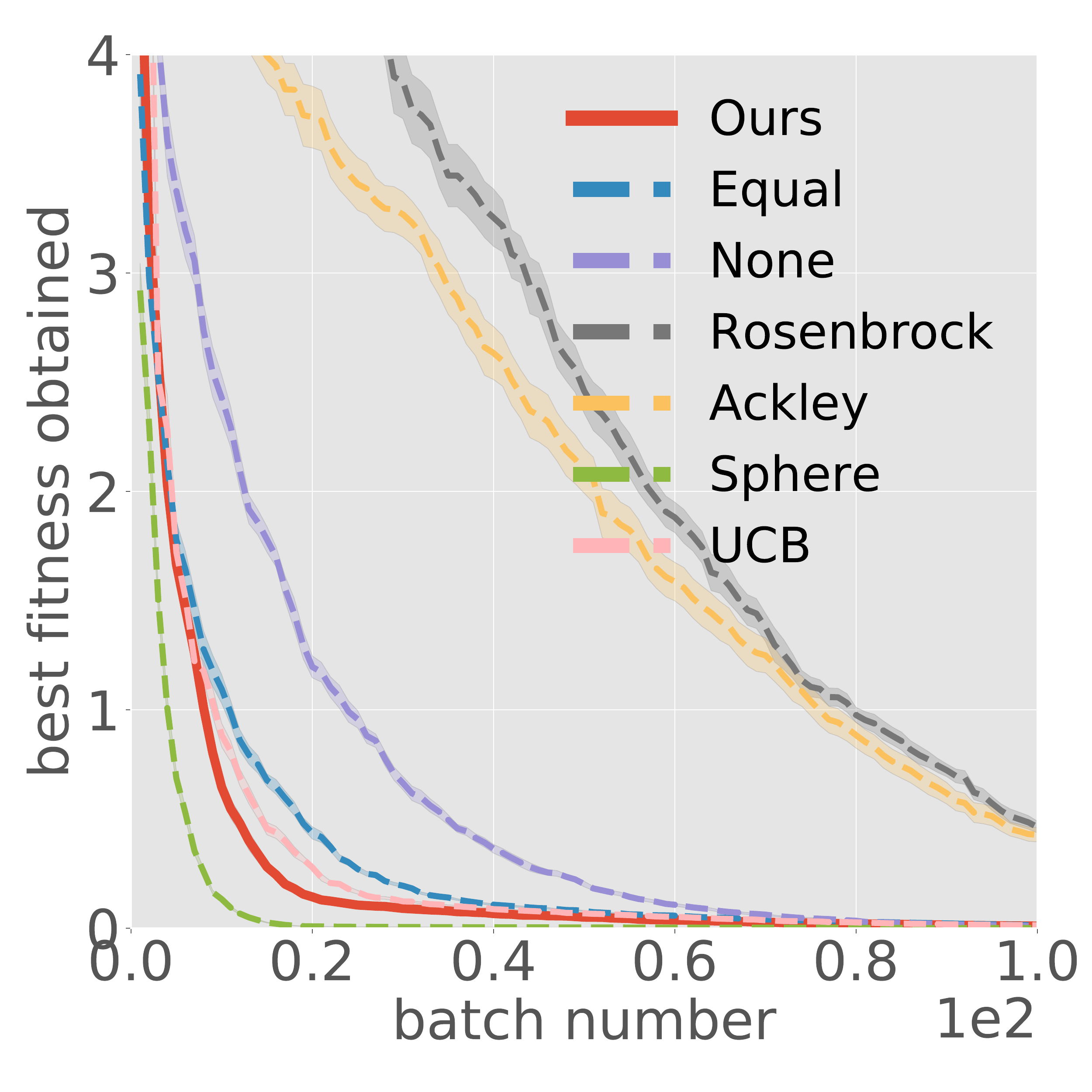}
        \includegraphics[width=0.99\textwidth]{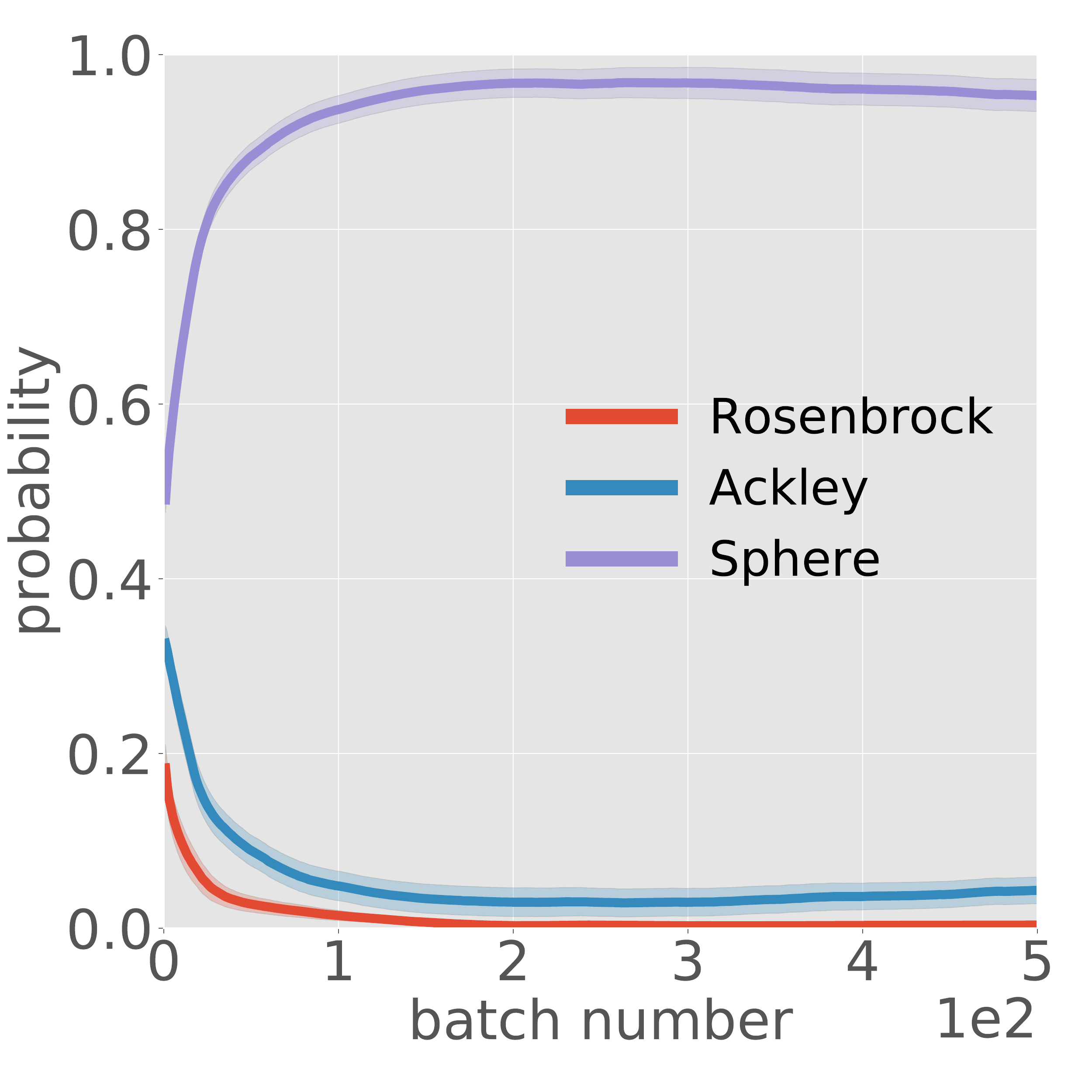}
        \caption{Sphere}
    \end{subfigure}%
    \begin{subfigure}{.195\textwidth}
        \centering
        \includegraphics[width=0.99\textwidth]{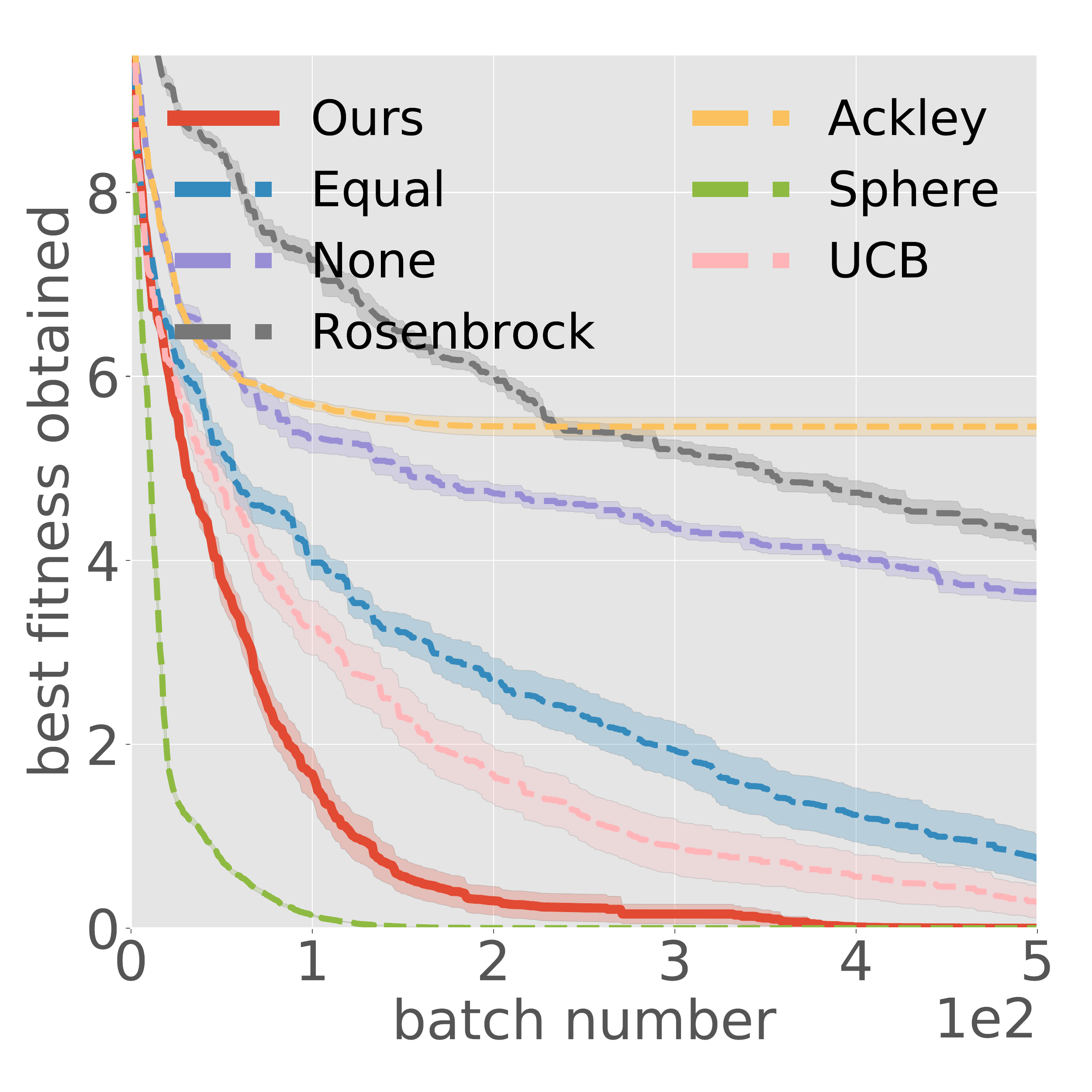}
        \includegraphics[width=0.99\textwidth]{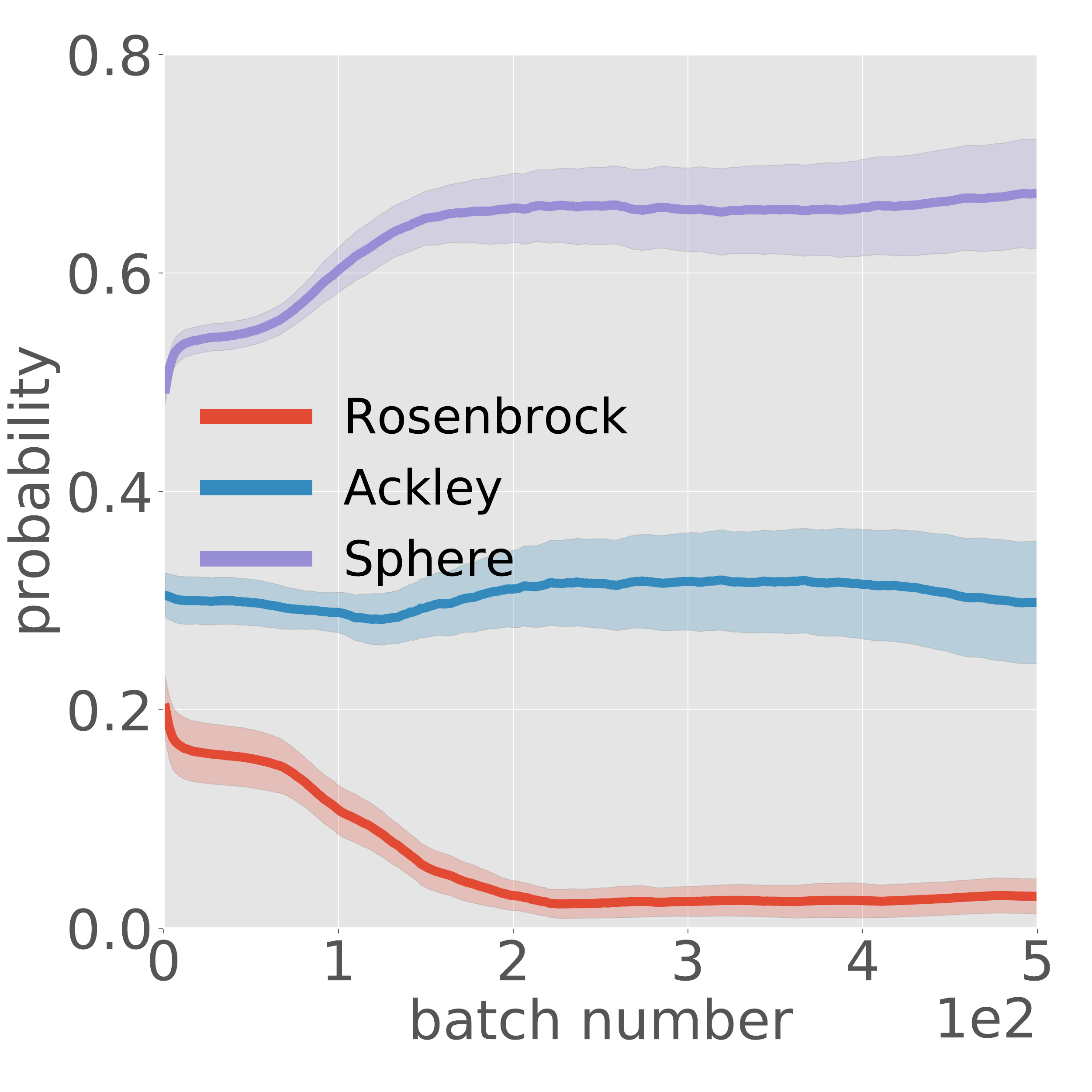}
        \caption{Rastrigin}
    \end{subfigure}%
    \begin{subfigure}{.195\textwidth}
        \centering
        \includegraphics[width=0.99\textwidth]{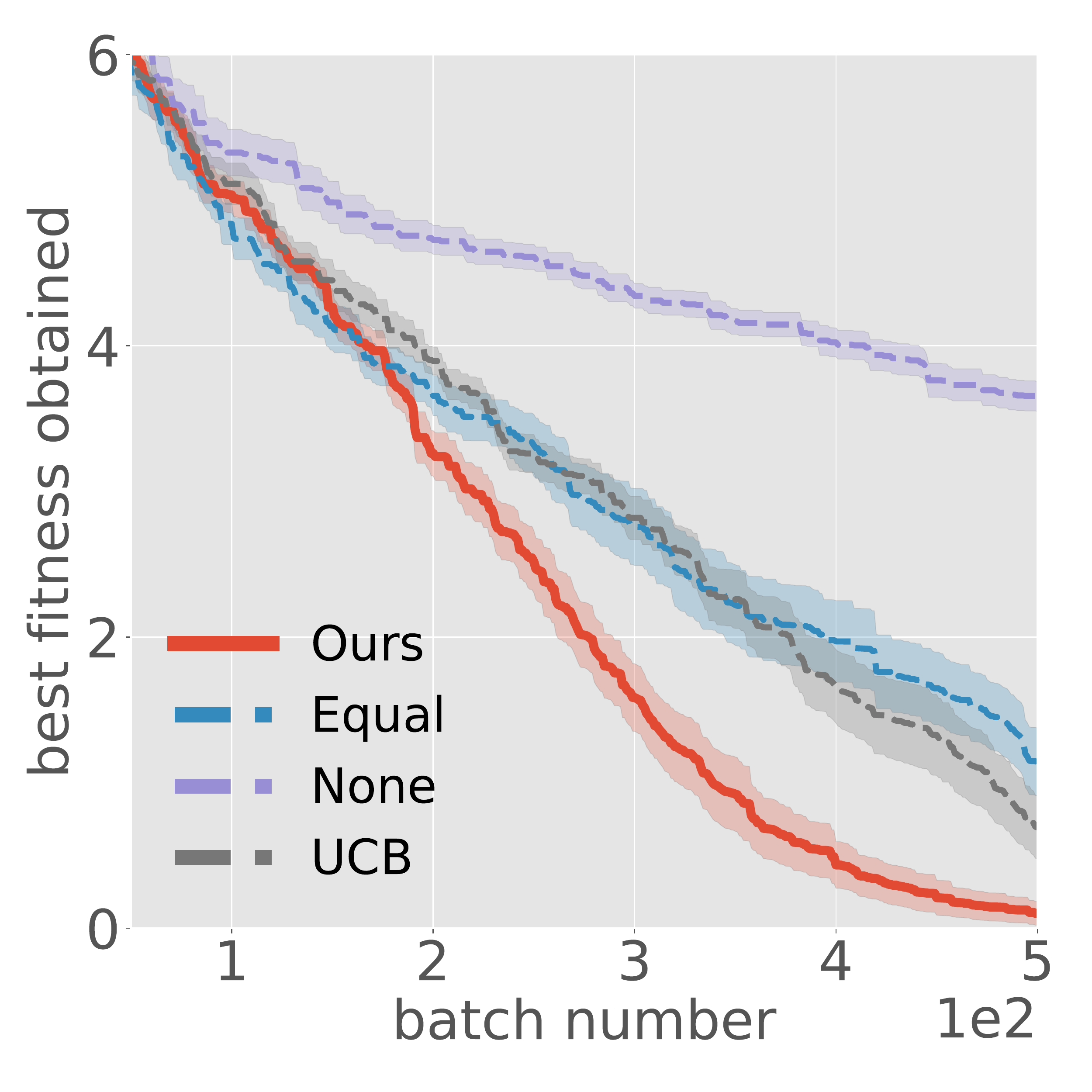}
        \includegraphics[width=0.99\textwidth]{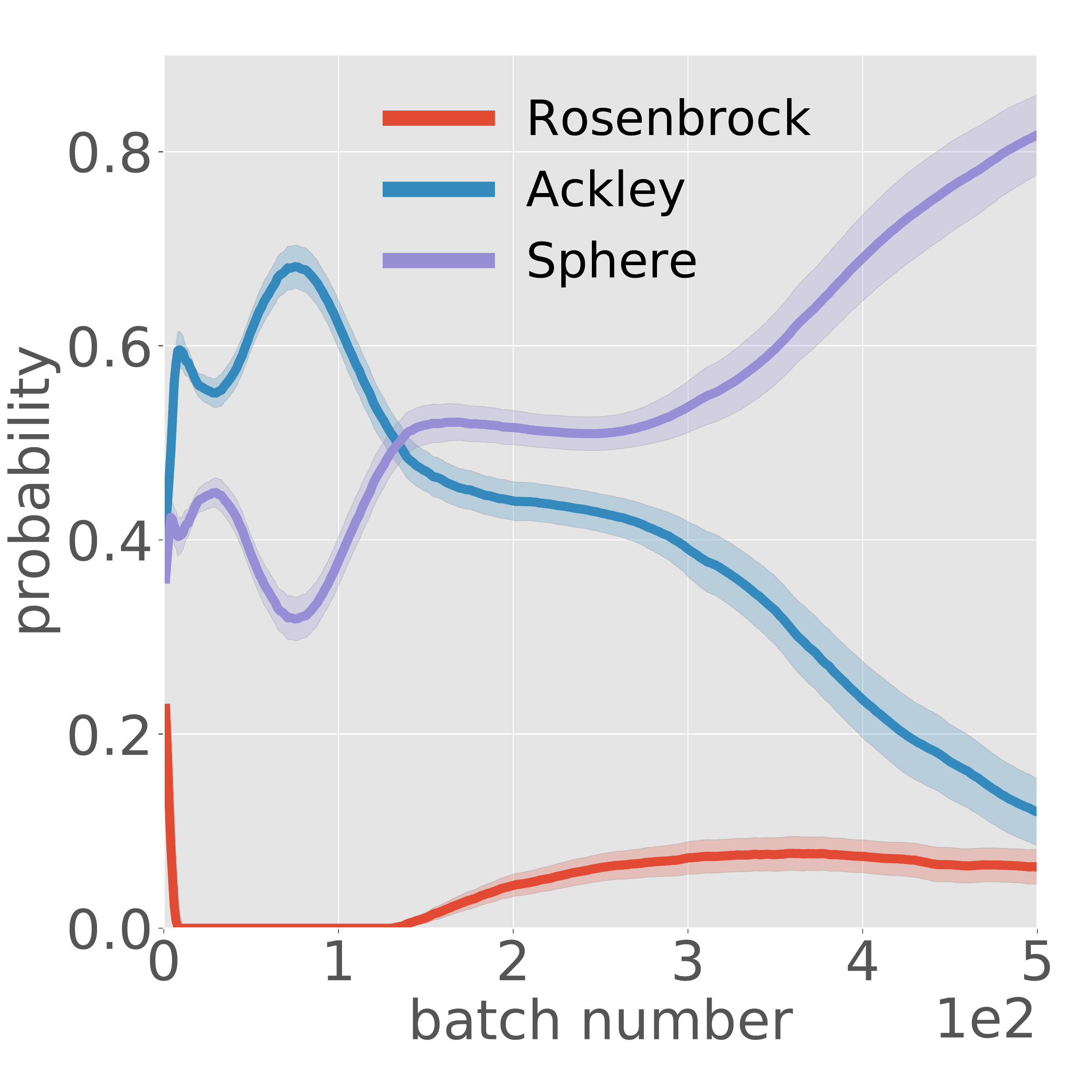}
        \caption{Rastrigin-MT}
    \end{subfigure}
    \caption{Best function fitness (top row) and weights assigned to source tasks (bottom row) over generations using transfer and multi-task (MT) learning for function optimization, with each source and target task as ground truth. Averaged over 20 trials with shaded standard error bars.}
    \label{fig:optimization_curve}
\end{figure}

\begin{figure}[!htb]
    \centering
    \begin{subfigure}{.245\textwidth}
        \centering
        \includegraphics[width=0.99\textwidth]{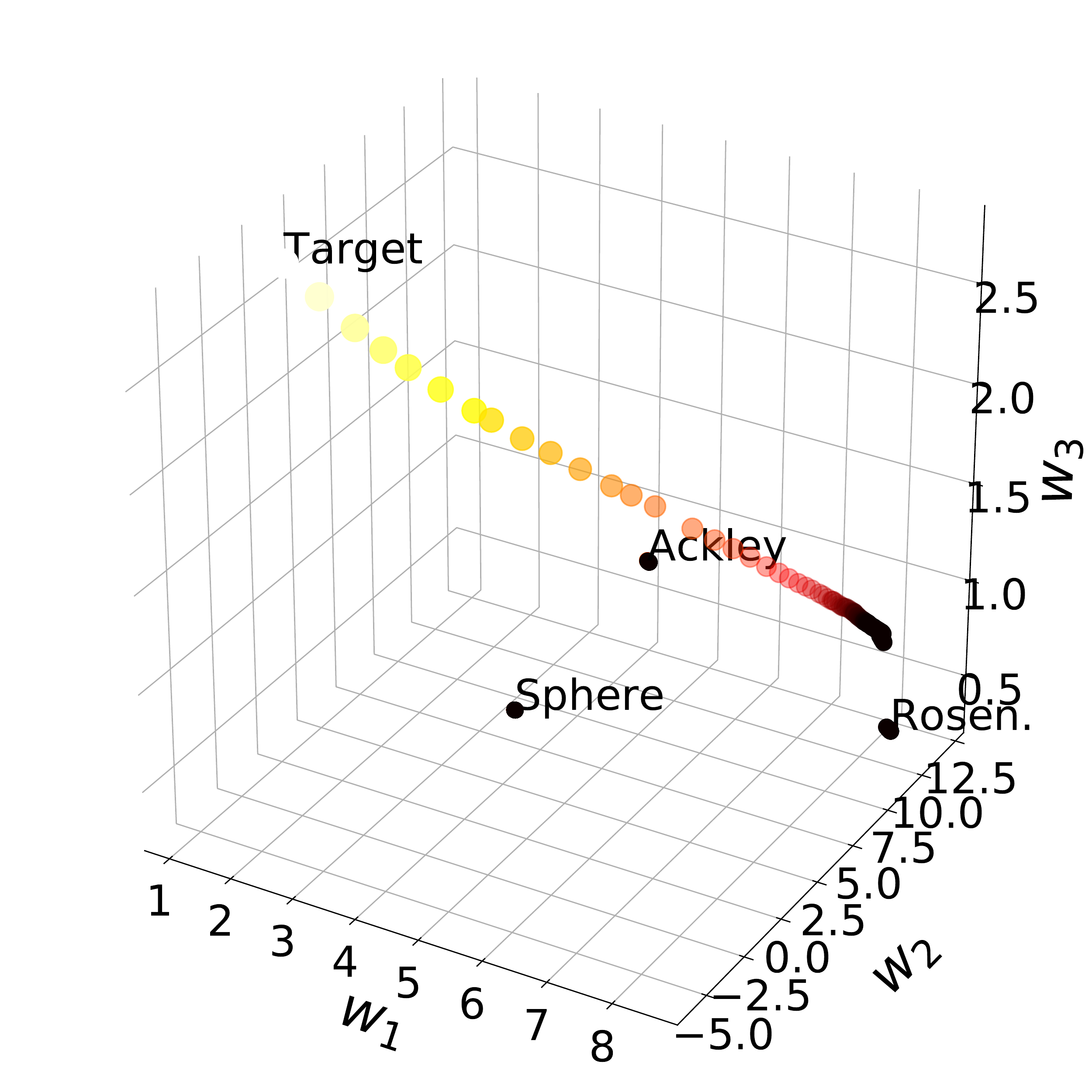}
        \caption{Rosenbrock}
    \end{subfigure}%
    \begin{subfigure}{.245\textwidth}
        \centering
        \includegraphics[width=0.99\textwidth]{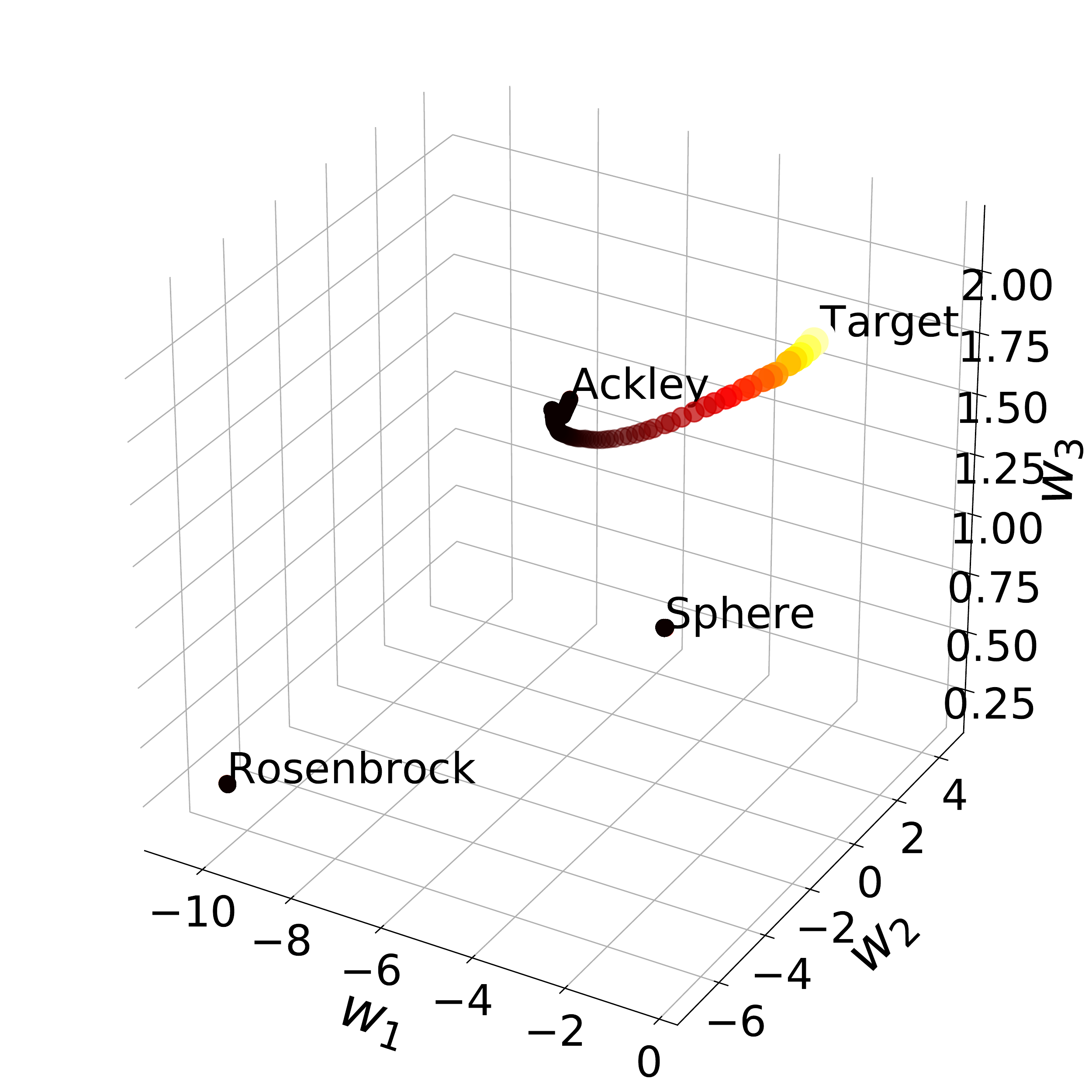}
        \caption{Ackley}
    \end{subfigure}%
    \begin{subfigure}{.245\textwidth}
        \centering
        \includegraphics[width=0.99\textwidth]{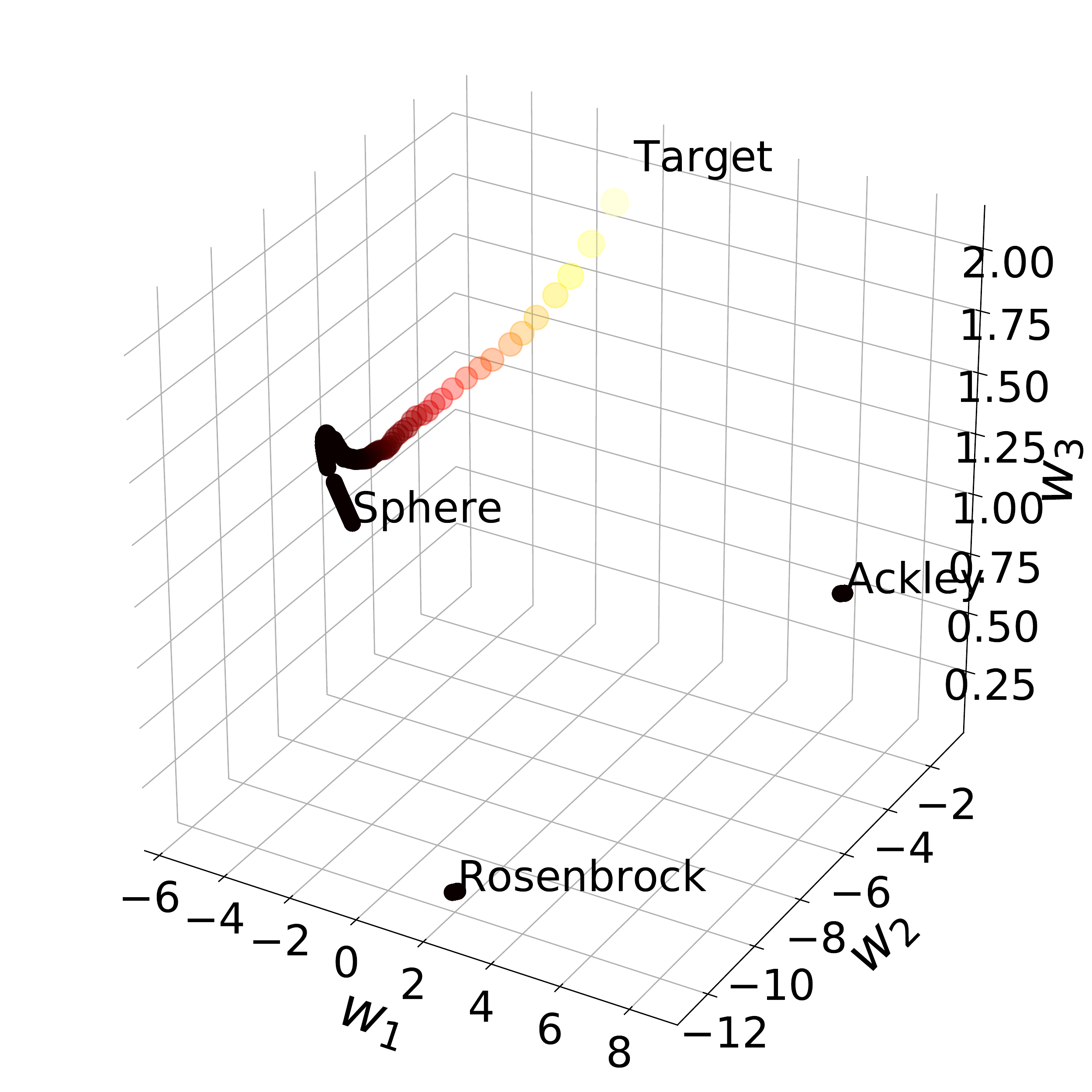}
        \caption{Sphere}
    \end{subfigure}%
    \begin{subfigure}{.245\textwidth}
        \centering
        \includegraphics[width=0.99\textwidth]{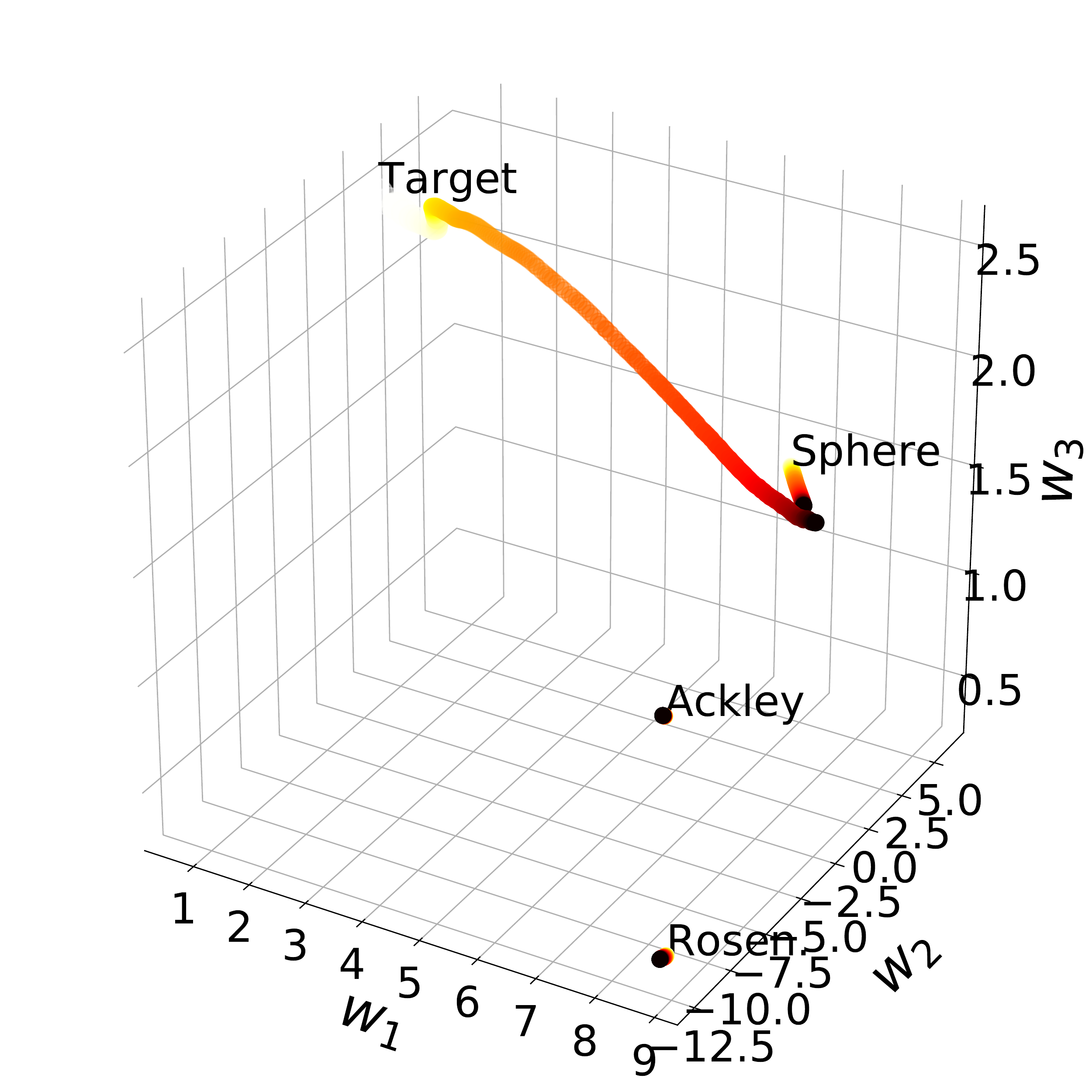}
        \caption{Rastrigin}
    \end{subfigure}
    \caption{For function optimization with each source and target task as the ground truth, shows the evolution of the posterior mean of each $\mathbf{w}_i$ and $\mathbf{w}_{target}$ during training for the simplified experiment with latent dimension $d = 2$. As shown here, the target mean eventually converges to the correct source task mean.}
    \label{fig:optimization_latent}
\end{figure}

We compare BERS (Ours) to the UCB algorithm \citep{auer2002using} with asymptotically optimal convergence ({UCB}), where the reward is improvement in best solution value between consecutive iterations. We also compare to the equal weighting of demonstrators ({Equal}), individual demonstrators (S1, S2\dots), and standard DE (None). Figure~\ref{fig:optimization_curve} illustrates the function value of the best solution and the QP solution in each iteration. Figure~\ref{fig:optimization_latent} tracks the movement of the posterior means of $\mathbf{w}_i$ and $\mathbf{w}_{target}$ during target training for a simplified model with $d = 2$ for visualization.

\subsection{Continuous Control of a Supply Chain Network}

\begin{figure}[!htb]
    \centering
    \begin{subfigure}{.3\textwidth}
        \begin{minipage}{0.49\linewidth}
            \includegraphics[width=0.99\textwidth]{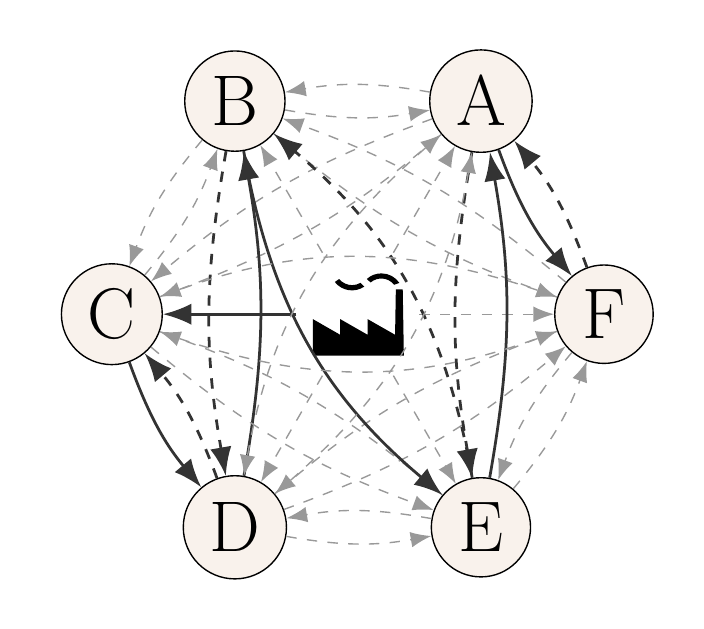}
            \caption*{Scenario 1}
        \end{minipage}
        \begin{minipage}{0.49\linewidth}
            \includegraphics[width=0.99\textwidth]{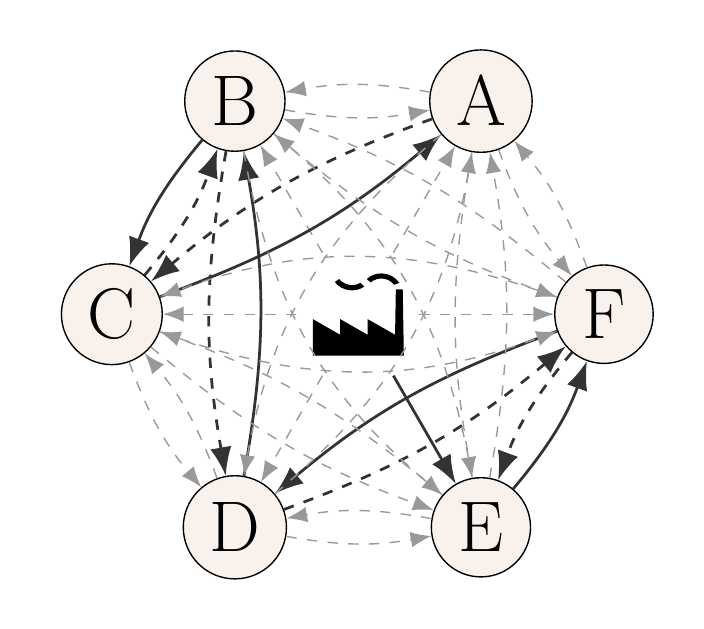} \caption*{Scenario 2}
        \end{minipage}
       \begin{minipage}{0.49\linewidth}
            \includegraphics[width=0.99\textwidth]{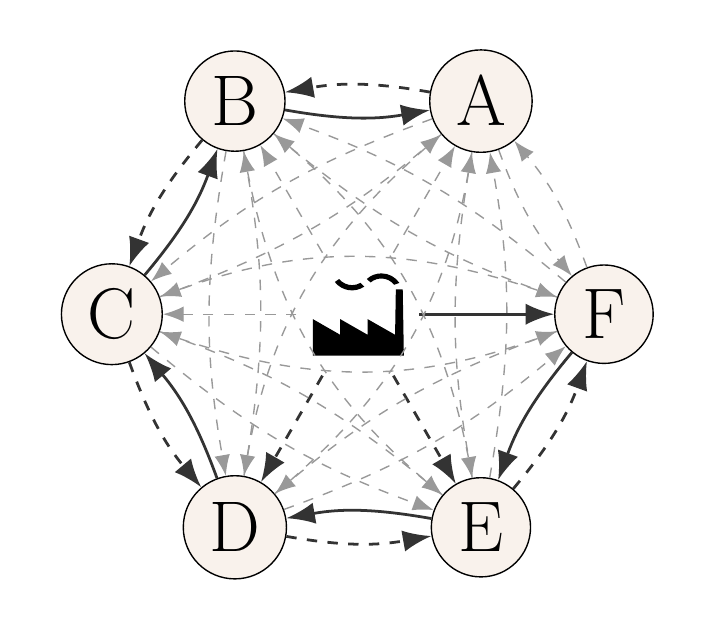}
            \caption*{Scenario 3}
        \end{minipage}
        \begin{minipage}{0.49\linewidth}
            \includegraphics[width=0.99\textwidth]{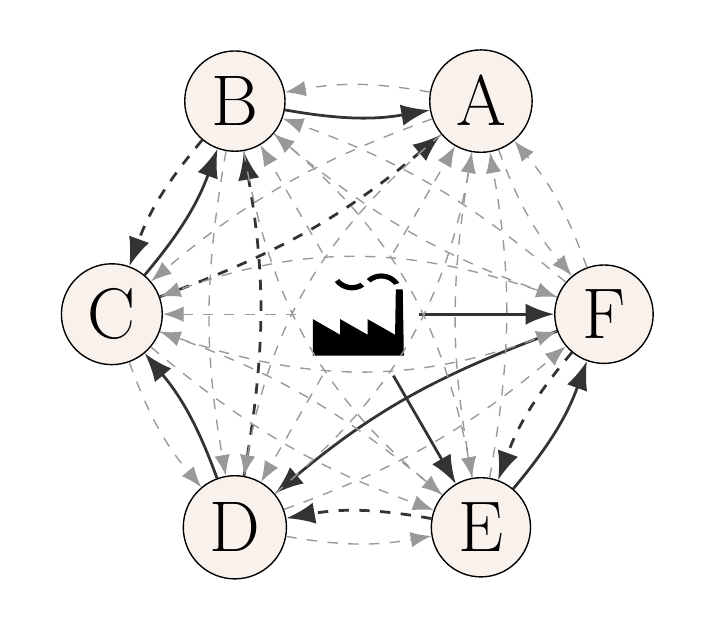}
            \caption*{Target}
        \end{minipage}
        \caption{Transport cost visualization.}
        \label{fig:source_target_supply}
    \end{subfigure}%
    \begin{subfigure}{0.7\textwidth}
        \includegraphics[width=0.99\linewidth]{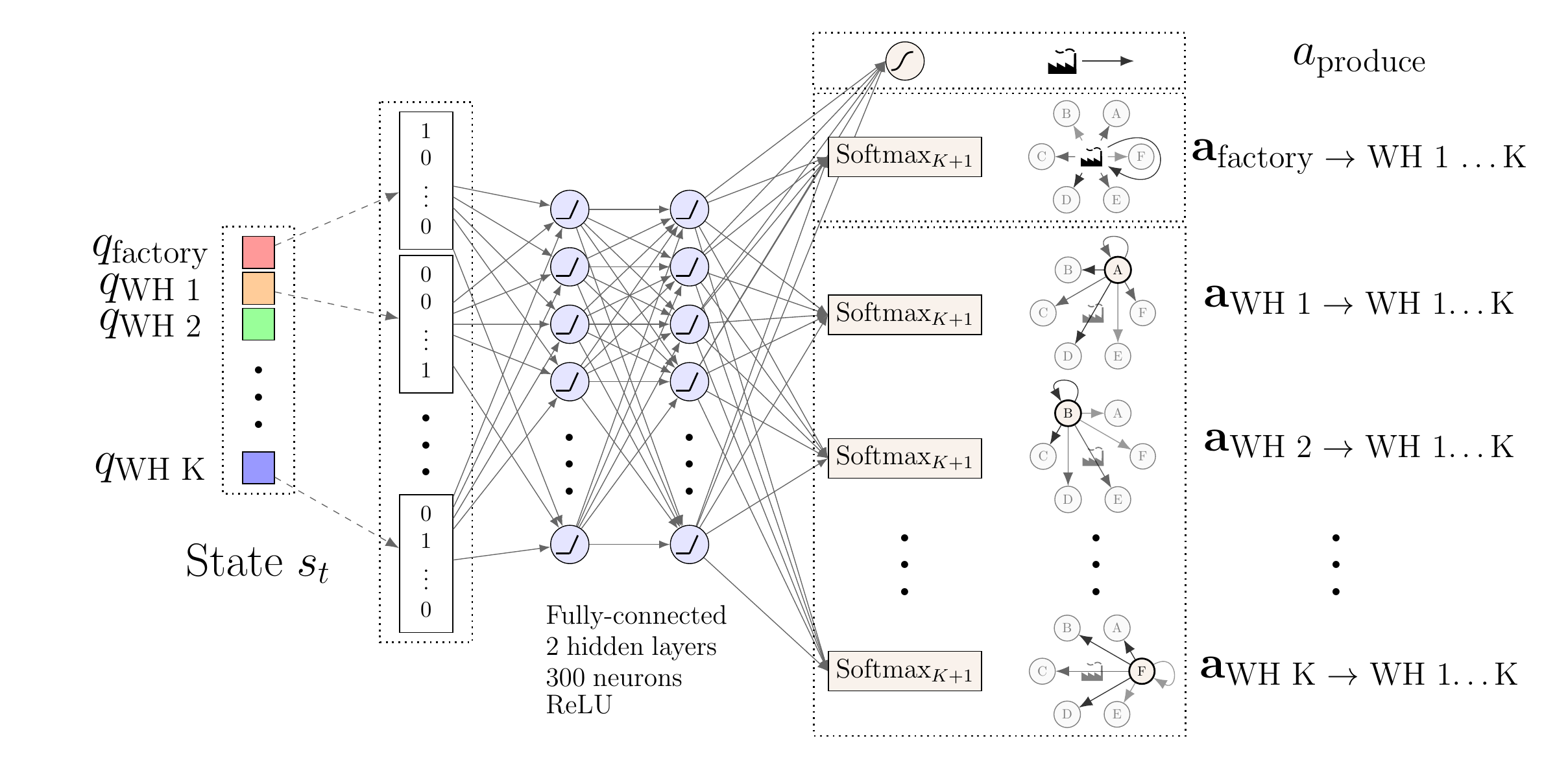} 
        \caption{Actor network.}
        \label{fig:supply_actor} 
    \end{subfigure} 
    \caption{Source and target task configuration and actor network used for the Supply Chain domain.}
    \label{fig:supply_network}
\end{figure}

The second problem is continuous control of a supply chain network as illustrated in Figure~\ref{fig:source_target_supply}. Here, the nodes include a central factory and $K = 6$ warehouses (denoted A, B \dots F), and edges represent possible transportation routes along which inventory can flow. A centralized controller must decide, at each discrete time step, how many units of a good to manufacture (up to 35 per period), and how many to ship from the factory to each warehouse and between warehouses. Furthermore, the cost of dispatching a truck depends on the source and destination. The controller identifies three likely scenarios (Scenarios 1, 2, and 3), shown in Figure~\ref{fig:source_target_supply}, where solid arrows indicate the cheapest routes with cost $0.03$ and dark and light grey arrows indicate costly routes with costs $1.50$ and $3.00$. A prudent agent must learn to take advantage of the ``cheapest" routes through the network while simultaneously learning optimal manufacturing and order quantities. 

The state includes the current stock in the factory and warehouses. Actions are modelled as follows: (1) one continuous action for production as a proportion of the maximum; (2) one set of $K + 1$ actions for proportions of factory stock to ship to each warehouse (including to keep at the factory); and (3) one set of $K$ actions per warehouse, for proportions of warehouse stocks to ship to all other warehouses (including itself). This leads to a $2 + K + K^2 = 44$-dimensional action space. In order to tractably solve this problem, we use the actor-critic algorithm DDPG \citep{lillicrap2015continuous} as the base learning agent $O_{base}$. The actor network is shown in Figure~\ref{fig:supply_actor}, using sparse encoding for states and sigmoid/ softmax output activation to satisfy stock constraints. The critic network has a similar structure.

\begin{figure}[!htb]
    \centering
    \begin{subfigure}{.5\textwidth}
        \centering
        \includegraphics[width=0.57\textwidth]{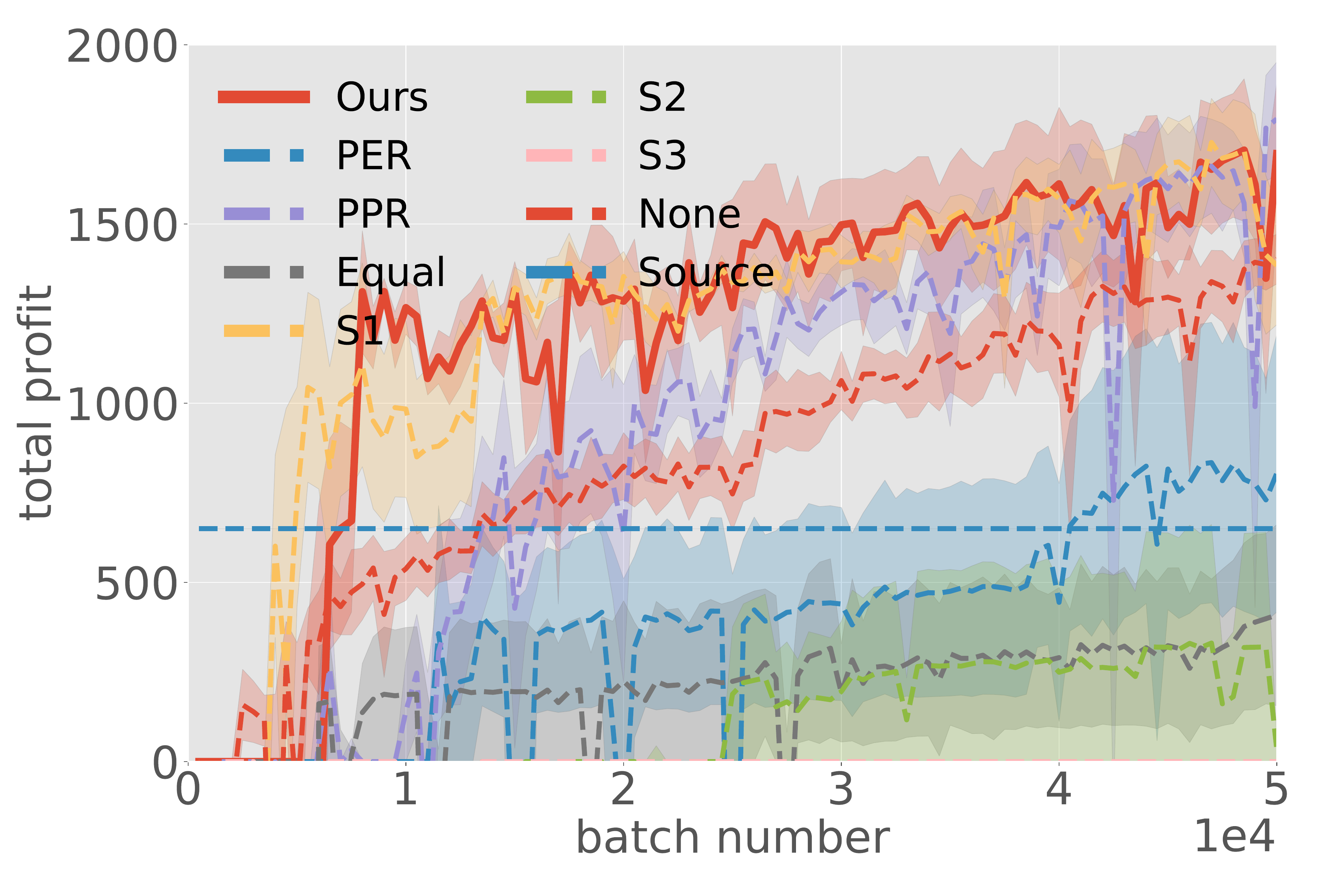}
        \includegraphics[width=0.3825\textwidth]{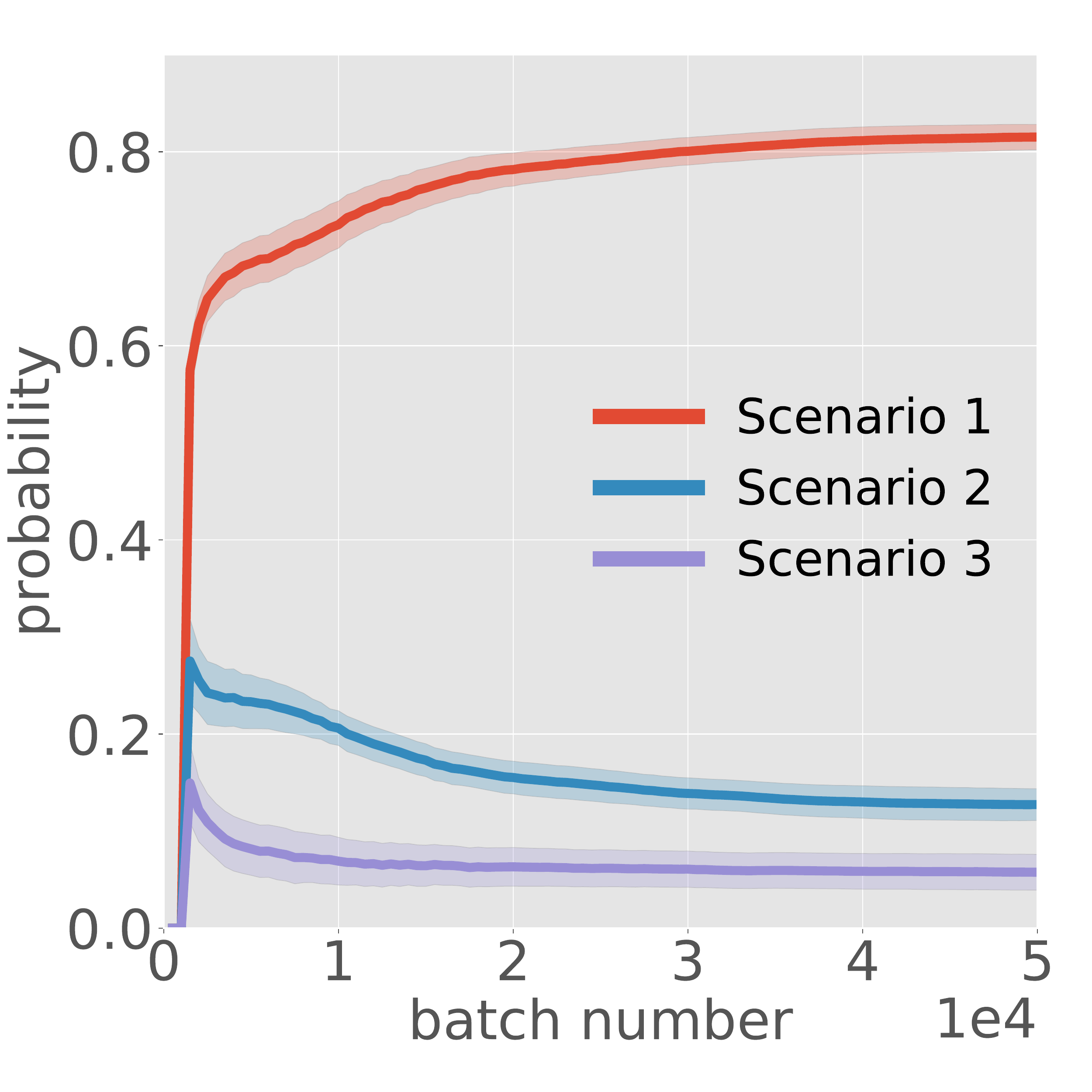}
        \caption{Scenario 1}
    \end{subfigure}%
    \begin{subfigure}{.5\textwidth}
        \centering
        \includegraphics[width=0.57\textwidth]{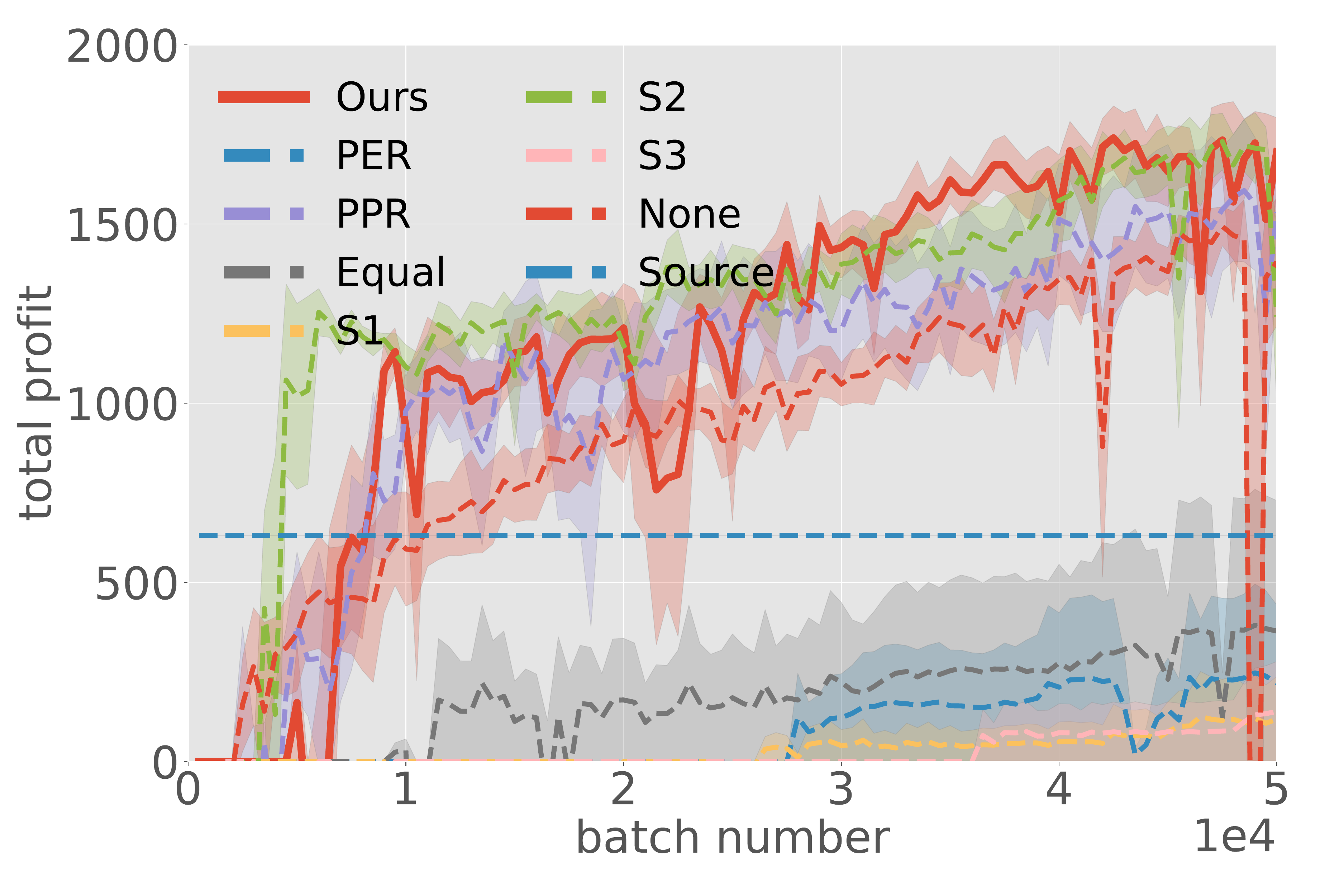}
        \includegraphics[width=0.3825\textwidth]{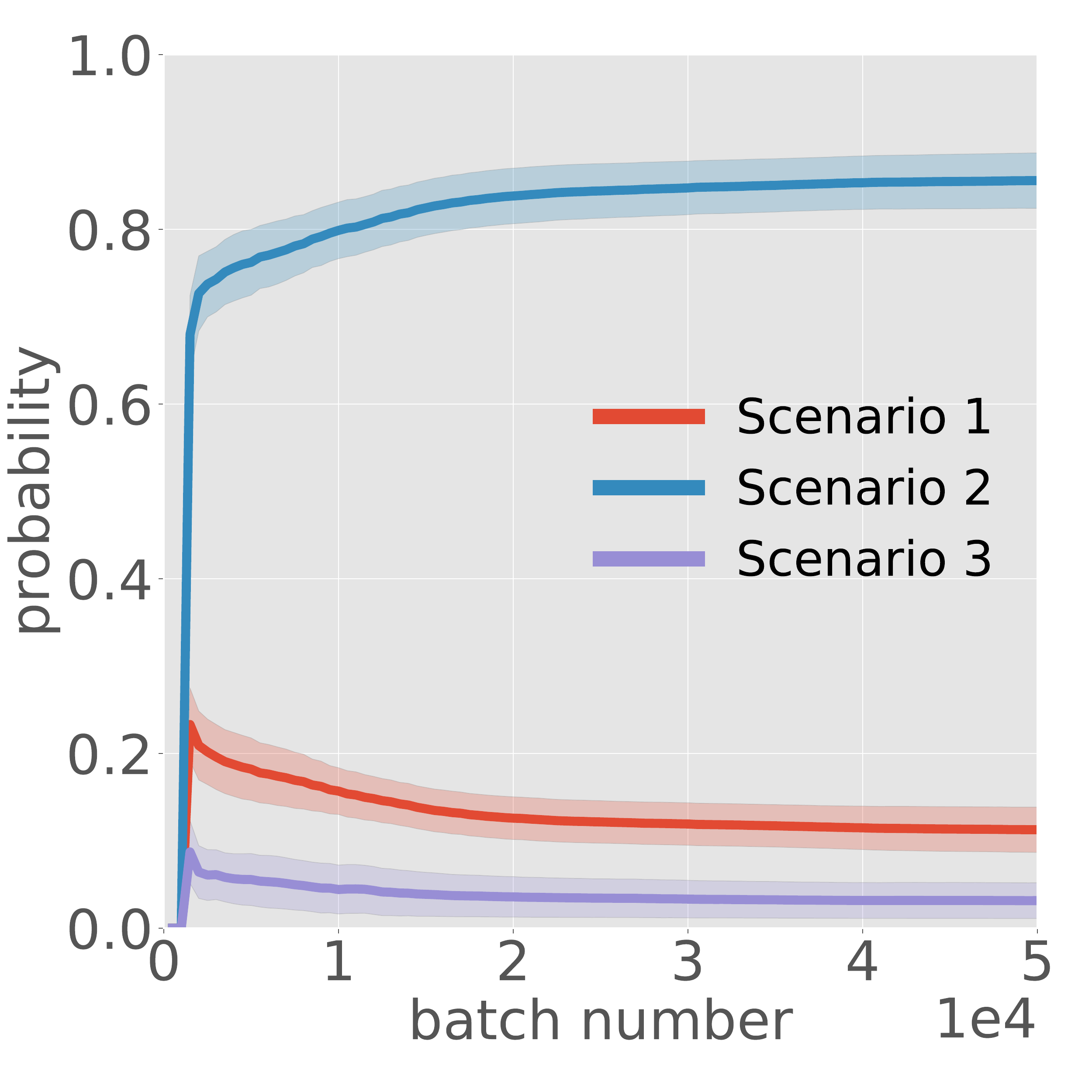}
        \caption{Scenario 2}
    \end{subfigure}
    \begin{subfigure}{.5\textwidth}
        \centering
        \includegraphics[width=0.57\textwidth]{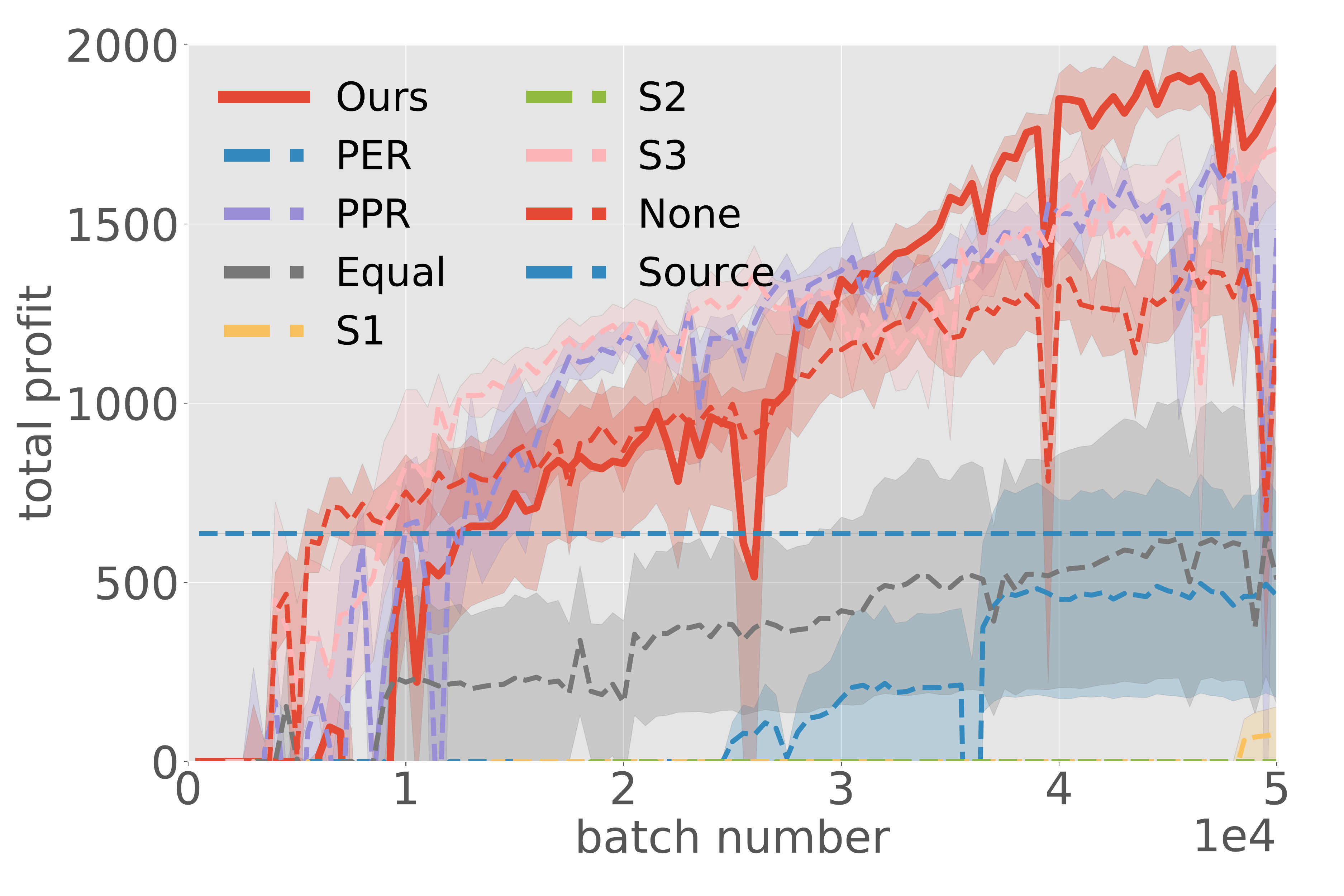}
        \includegraphics[width=0.3825\textwidth]{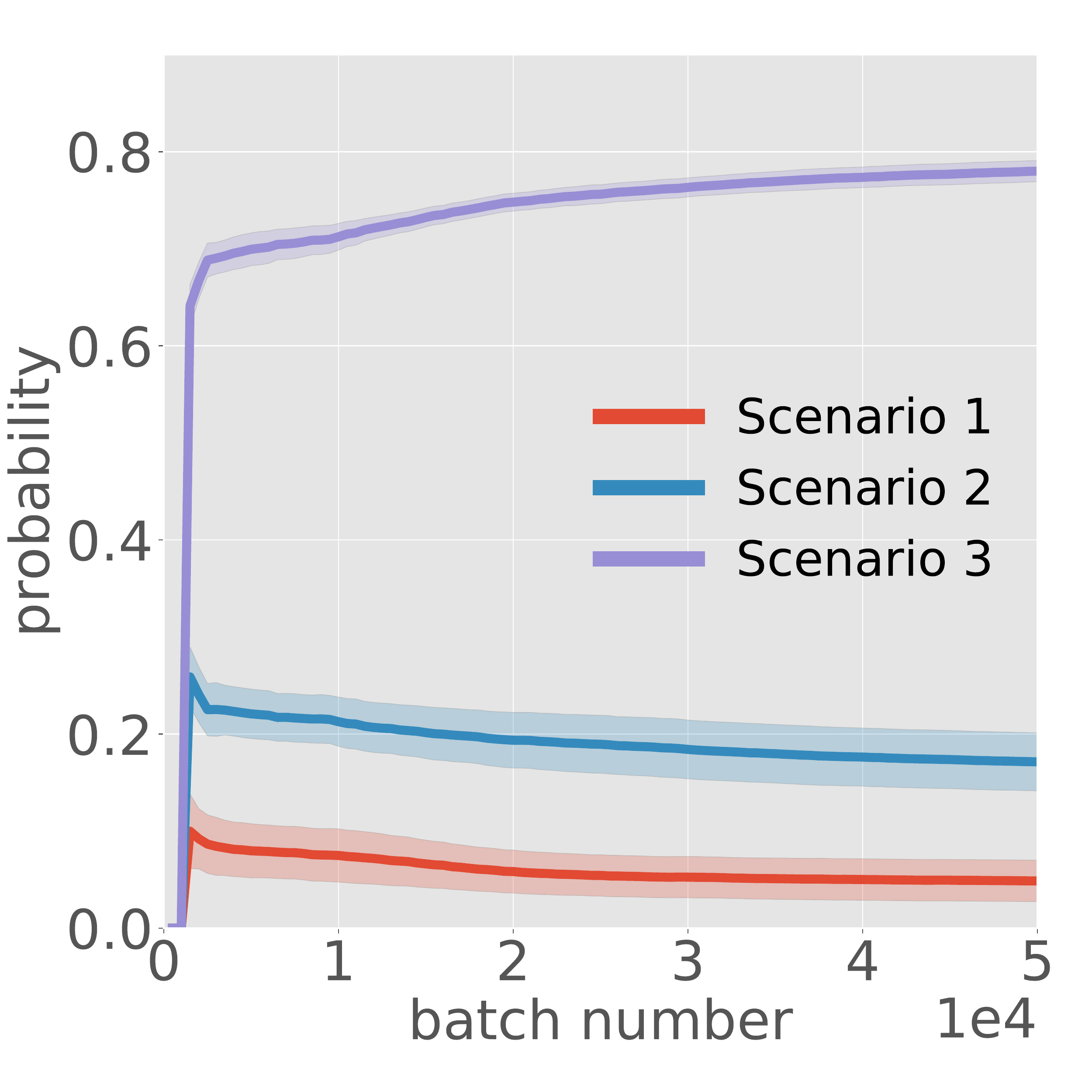}
        \caption{Scenario 3}
    \end{subfigure}%
    \begin{subfigure}{.5\textwidth}
        \centering
        \includegraphics[width=0.57\textwidth]{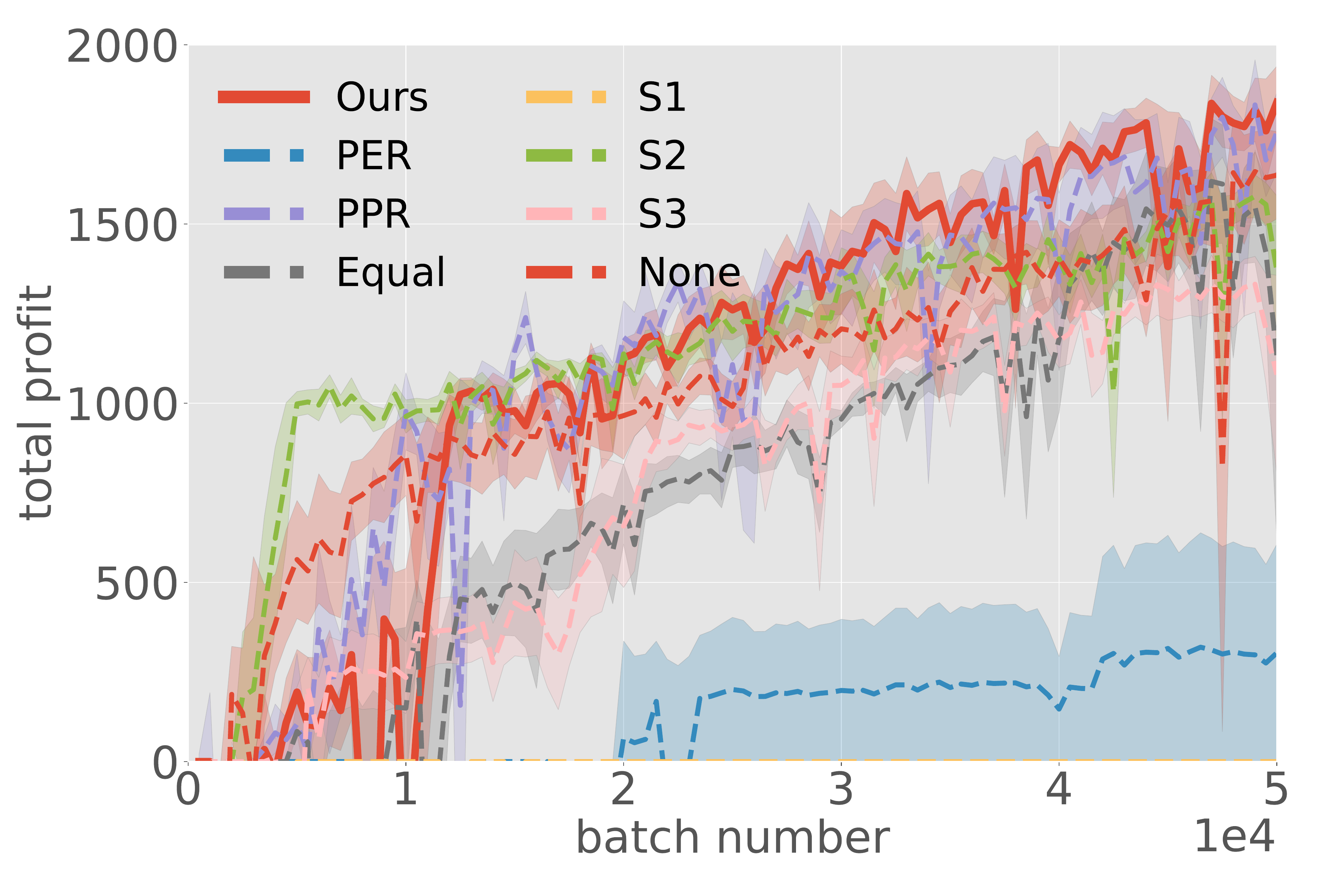}
        \includegraphics[width=0.3825\textwidth]{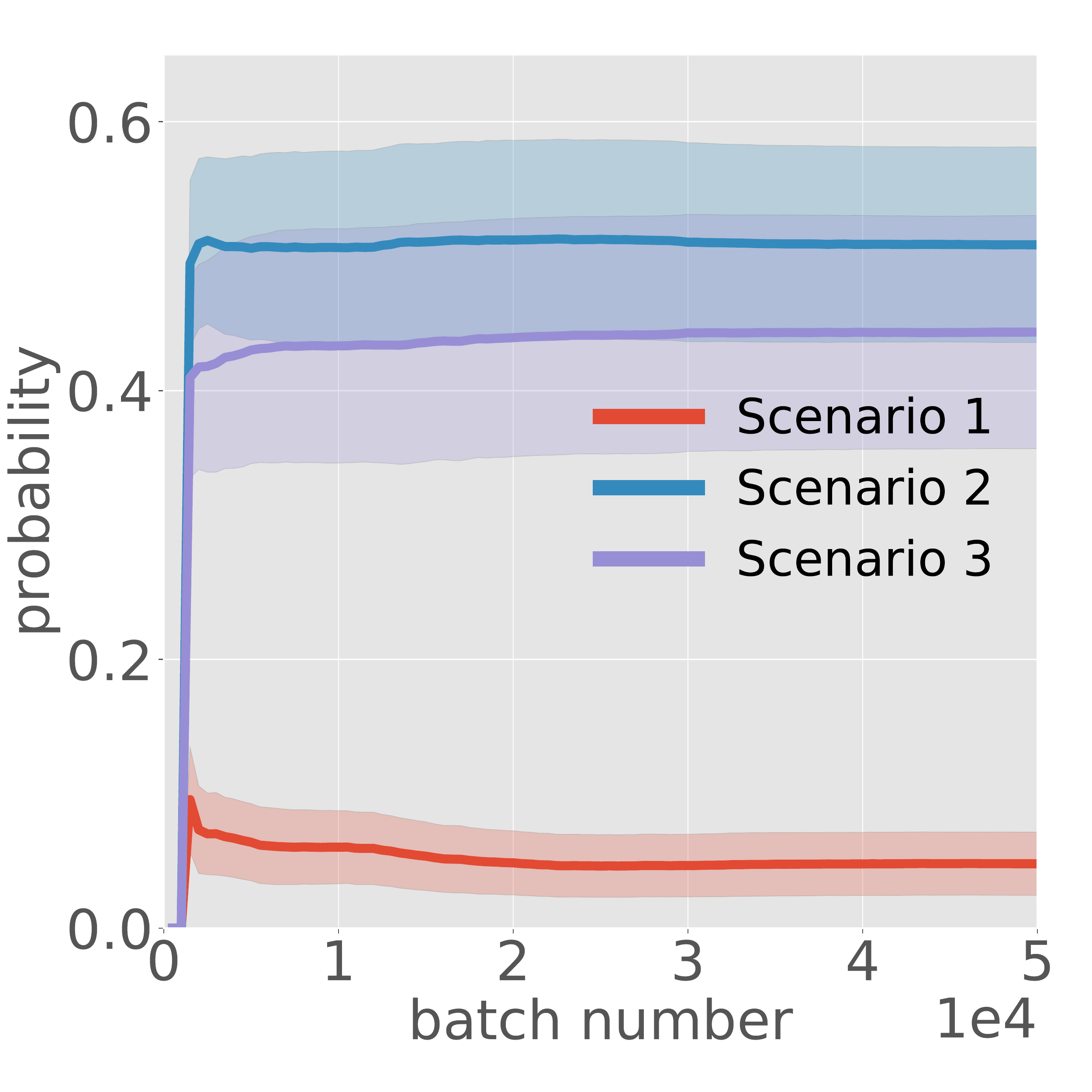}
        \caption{Target Task}
    \end{subfigure}
    \caption{Total testing profit per episode (left) and weights assigned to source tasks (right) over epochs using soft pre-training of DDPG for the Supply Chain problem, with each source and target task as ground truth. Averaged over 5 trials with shaded standard error bars.}
    \label{fig:supply_curve}
\end{figure}

\begin{figure}[!htb]
    \centering
    \begin{subfigure}{.245\textwidth}
        \centering
        \includegraphics[width=0.95\textwidth]{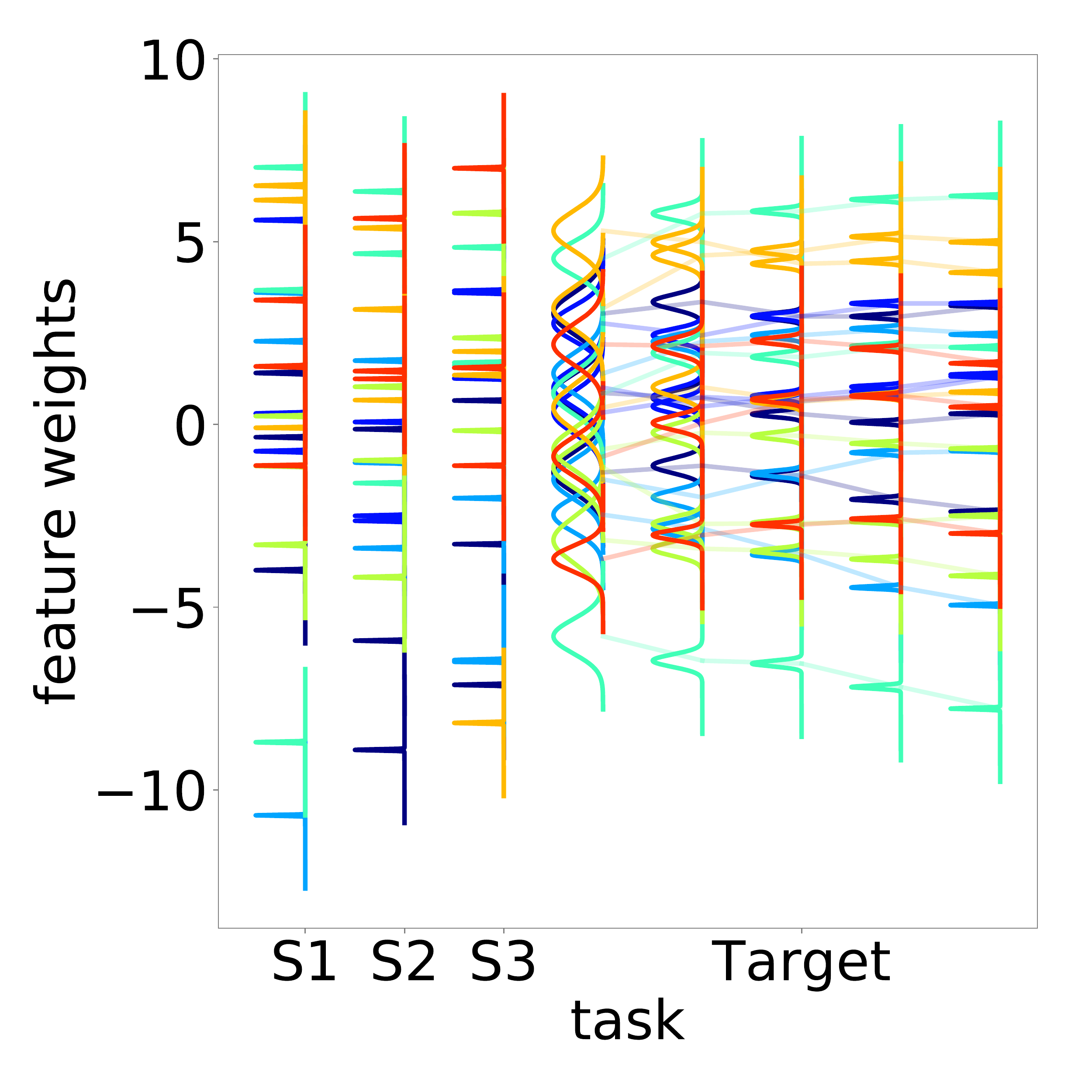}
        \caption{Scenario 1}
    \end{subfigure}%
    \begin{subfigure}{.245\textwidth}
        \centering
        \includegraphics[width=0.95\textwidth]{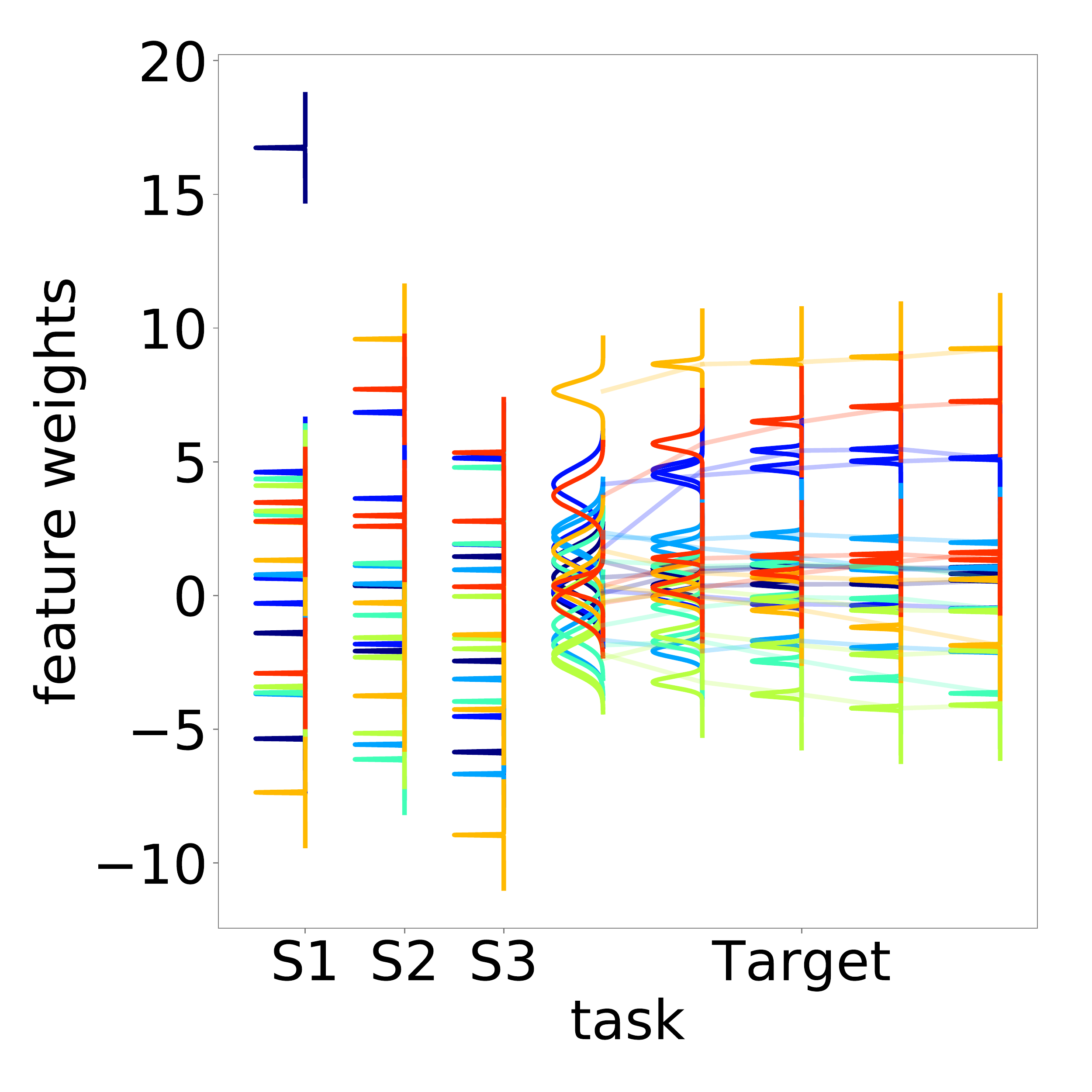}
        \caption{Scenario 2}
    \end{subfigure}%
    \begin{subfigure}{.245\textwidth}
        \centering
        \includegraphics[width=0.95\textwidth]{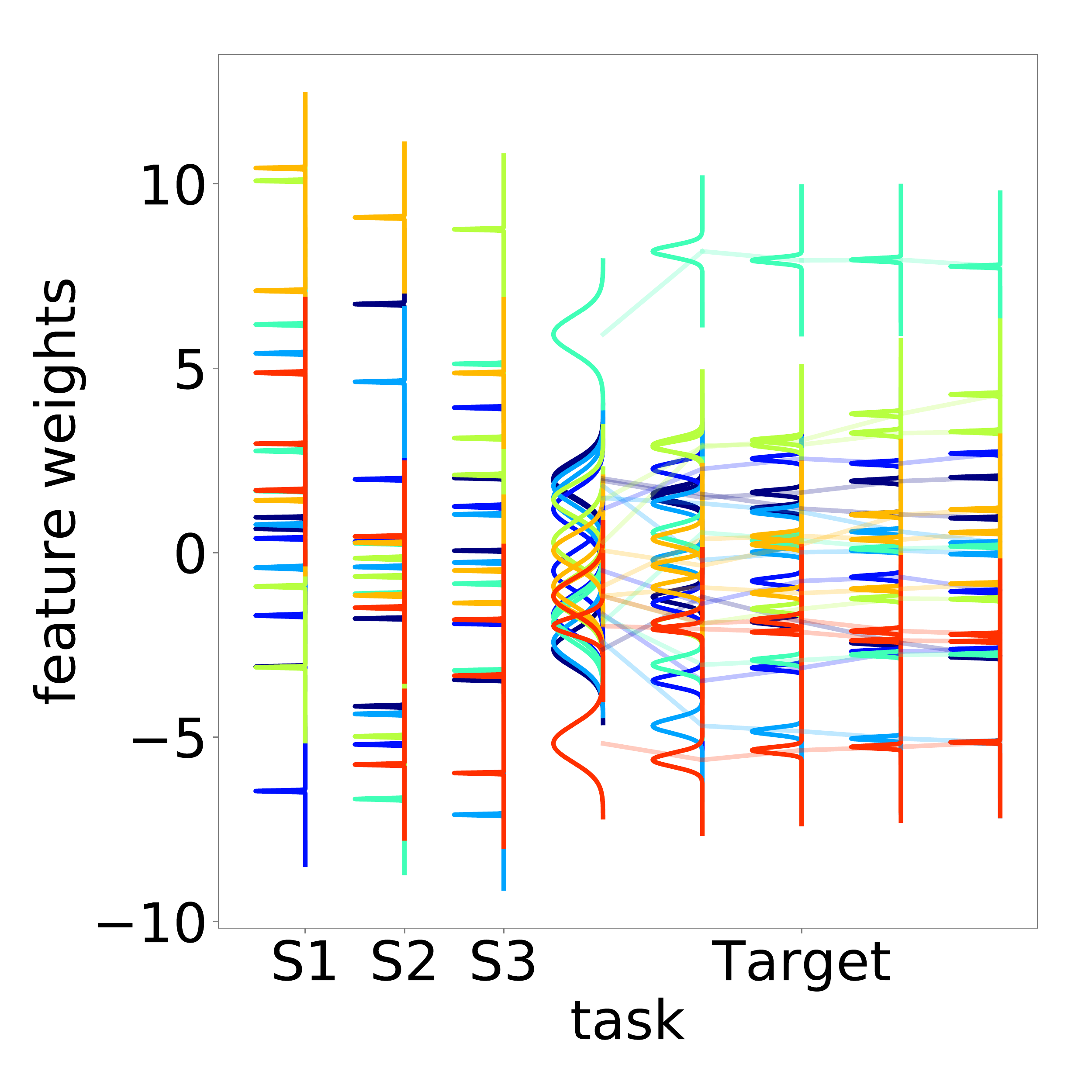}
        \caption{Scenario 3}
    \end{subfigure}%
    \begin{subfigure}{.245\textwidth}
        \centering
        \includegraphics[width=0.95\textwidth]{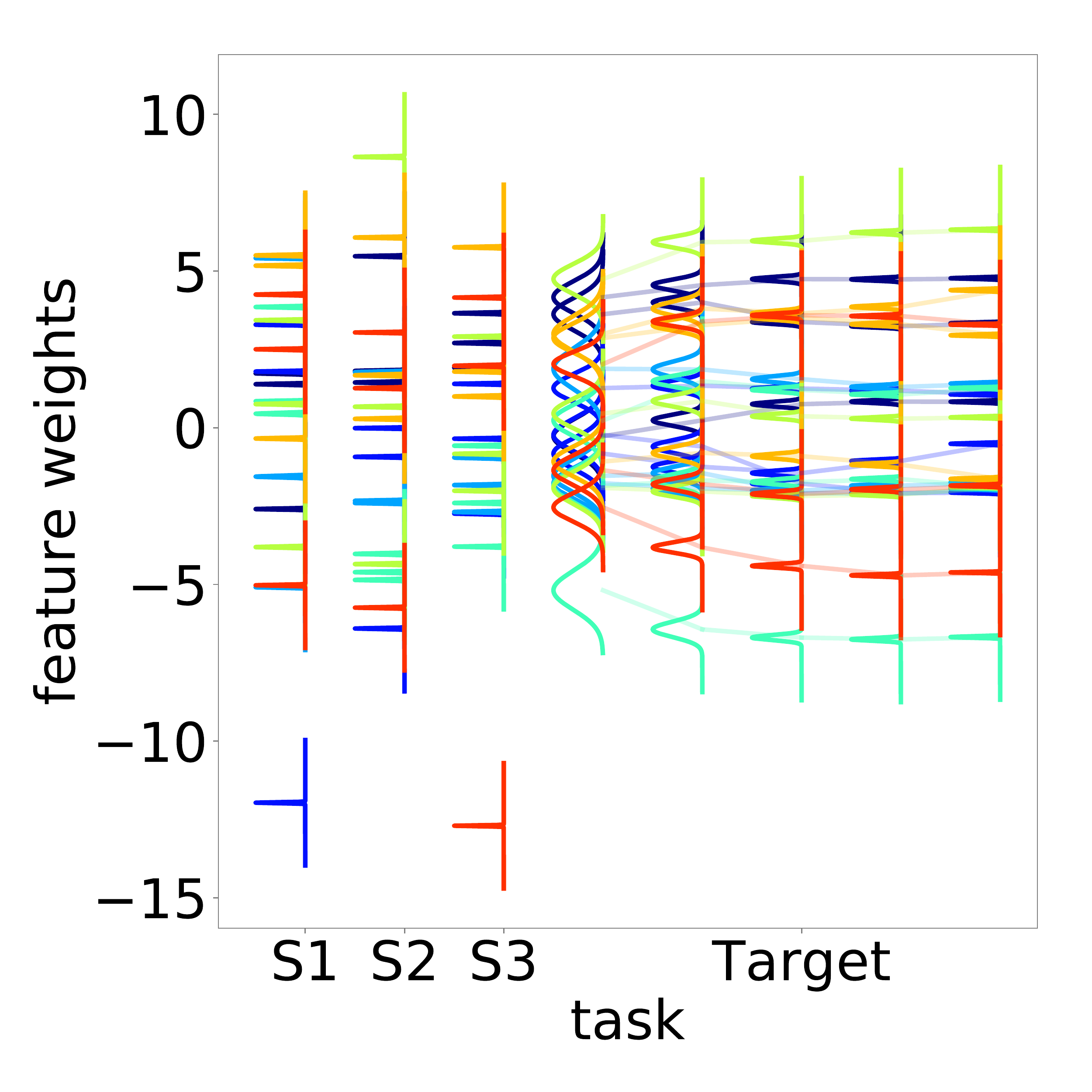}
        \caption{Target}
    \end{subfigure}
    \caption{For Supply Chain control with each source and target task as the ground truth, shows the posterior marginal distributions of $\mathbf{w}_i$ and $\mathbf{w}_{target}$ during training on the target task (after 0, 10, 50, 100 and 200 episodes). Over time, the target features begin to concentrate on the corresponding values for the correct source task.}
    \label{fig:supply_latent}
\end{figure}

We evaluate BERS (Ours) against prioritized experience replay \citep{hou2017novel,schaul2015prioritized} with pre-loaded demonstrations from all source tasks (PER), and HAT \citep{taylor2011integrating} combined with a state-of-the-art policy reuse algorithm \citep{li2018optimal} (PPR). In the latter case, source policies are trained using the same architecture as the actor network in Figure~\ref{fig:supply_actor} for 50 epochs using a cross-entropy loss. Figure~\ref{fig:supply_curve} illustrates the total test profit obtained, as well as the weights assigned by the QP problems to the source tasks. Figure~\ref{fig:supply_latent} how the posterior weights of the target task approach the weights of the ground truth.

\subsection{Discussion}

On the optimization task, BERS achieves a small performance gap relative to the single best expert, since it can quickly identity the most suitable source task (Figure~\ref{fig:optimization_curve}). While UCB is a strong baseline, BERS finds the solution in less iterations and with less variability. As expected, the QP solution favours the Sphere function as the source task when solving the Rastrigin function, because they are structurally similar functions, which leads to a quicker identification of the global minimum.

On the supply chain task, BERS also achieves results that are similar to the single best expert, and does better than PPR (while the asymptotic performance on Scenario 1 is similar, BERS achieves better jump-start performance). Both BERS and PPR are able to quickly surpass the performance of the exploration policy that generated the source data (shown as a horizontal line in Figure~\ref{fig:supply_curve}), whereas PER was not able to obtain satisfactory performance. We believe the latter is true because PER prioritizes experiences by TD error, which is not suitable when the rewards are sampled from different tasks, while BERS learns a common feature embedding to allow consistent comparison between tasks (Figure~\ref{fig:optimization_latent}). We also believe that if the source data was considerably less optimal, then the performance gap between PPR and BERS would be greater. On the target scenario, it is interesting to see that the weights assigned to Scenarios 1 and 2 are roughly equal, which makes sense as the target task shares similarities with both scenarios (Figure~\ref{fig:source_target_supply}). By mixing two source tasks, BERS was able to perform substantially better on the target task than the two source tasks (S2 and S3) in isolation. Also by adopting a fully Bayesian treatment, we obtain smooth convergence of the task weights on all experiments.

\section{Conclusion}

We studied the problem of LfD with multiple sub-optimal demonstrators with different goals. To solve this problem, we proposed a multi-headed Bayesian neural network with a shared encoder to efficiently learn consistent representations of the source and target reward functions from the demonstrations. Reward functions were parameterized as linear models, whose uncertainty was modeled using Normal-Inverse-Gamma priors and updated using Bayes' rule. A quadratic programming problem was formulated to rank the demonstrators while trading off the mean and variance of the uncertainty in the learned reward representations, and Bayesian Experience Reuse (BERS) was proposed to incorporate demonstrations directly when learning new tasks. Empirical results for static function optimization and RL suggest that our approach can successfully transfer experience from conflicting demonstrators in diverse tasks.

\section*{Broader Impact}

This work presents a general framework for reusing demonstrations data available in general static and dynamic optimization problems. It could lead to positive societal consequences in terms of safety and reliability of automated systems, such as medical diagnosis systems that often require pre-training from human experts, or driver-less cars that learn from human mistakes to avoid similar situations in the real world, potentially saving lives in the process. On the other hand, because our approach requires the learning of the (latent) intentions of users and their goals, there could potentially be privacy concerns depending on how the trained agent decisions are implemented or published. This is particularly true shortly after commencement of training, when the learning agent can resemble one or more demonstrators in their decision making. Our framework is also based on deep learning methodologies, and so it naturally inherits some of the risks prevalent in that work (although our work also tries to address some of these issues by using a Bayesian approach). For example, data that consists of only exceptional situations (such as pre-training of a search and rescue robot), may consist of many outliers. Such data sets must sometimes undergo pre-processing to remove the effects of outliers or other features that could skew the algorithm and lead to the wrong conclusions. This could be catastrophic in mission critical operations where every decision counts. However, by using prudent machine learning practices, we believe the benefits of this work outweigh the risks.

\begin{ack}
This work was funded by a DiDi Graduate Student Award. The authors would like to thank Jihwan Jeong for his Python code for training the neural-linear network and discussions that helped improve this paper.  
\end{ack}

\small

\bibliography{main.bib}
\bibliographystyle{plainnat}

\newpage
\normalsize
\setcounter{page}{1}

\section*{Appendix}

\subsection*{Hyper-Parameters for the Neural-Linear Model}

We set $\mathbf{w}_i \sim \mathcal{N}(0, \mathbf{I})$, $\sigma_i^2 \sim \mathrm{InvGamma}(1, 1)$. For the encoder, we set $d = 20$, use L2 regularization with penalty $\lambda = 10^{-4}$, a learning rate of $10^{-4}$, and initialize weights using Glorot uniform. The encoder has two hidden layers with 200 ReLU units each (300 in the first layer for the supply chain problem), and $\tanh$ outputs. We also include a constant bias term in the feature map $\bm{\phi}$ when learning $\mathbf{w}$. Prior to transfer, we first train the neural linear model on 4000 batches of size 64 sampled uniformly from source data. During target training, we train on batches of size 64 sampled from source and target data after each iteration (generation for optimization, episode for supply chain) to avoid catastrophic forgetting of features (one batch for optimization and 20 for supply chain). QPs are solved from scratch at the end of each iteration using the \texttt{cvxopt} package \citep{andersen2015cvxopt}.

\subsection*{Details for the Static Optimization Benchmark}

\paragraph{Problem Setting:} The four test functions considered in this paper are as follows:
\begin{align*}
    f_{Rosenbrock}(\mathbf{x}) &= \sum_{i=1}^{D - 1} 100 [(x_{i+1} - x_i^2)^2 + (1 - x_i)^2] \\
    f_{Ackley}(\mathbf{x}) &= -20 \exp\left(-0.2 \sqrt{\frac{1}{D} \sum_{i=1}^D x_i^2} \right) - \exp\left(\frac{1}{D} \sum_{i=1}^D \cos(2 \pi x_i) \right) + 20 + \exp{(1)} \\
    f_{Sphere}(\mathbf{x}) &= \sum_{i=1}^D (x_i+2)^2 \\
    f_{Rastrigin} &= 10 D + \sum_{i=1}^D [(x_i+2)^2 - 10 \cos(2 \pi (x_i+2))],
\end{align*}
where $\mathbf{x} = [x_1, x_2 \dots x_D] \in \mathbb{R}^D$ and restricted to $x_i \in [-4, 4]$ for $i = 1, 2 \dots D$ and we set $D = 10$ for all experiments. The global minimums of the functions are, respectively, as follows: $x_{Rosenbrock}^* = \bm{1}_D$, $x_{Ackley}^* =  \bm{0}_D$, $x_{Sphere}^* = -2 \times \bm{1}_D$ and $x_{Rastrigin}^* = -2 \times \bm{1}_D$. For Rosenbrock, Sphere and Rastrigin functions, we transform outputs using $y \mapsto \sqrt{y}$ so they become, approximately, equally scaled (for Rosenbrock we subsequently also divide by 10), to demonstrate whether we can actually ``learn" each function from the data rather than distinguish it according to scale alone. 

\paragraph{Solver Settings:} To optimize all functions, we use the Differential Evolution (DE) algorithm~\citep{storn1997differential}. The pseudo-code of this algorithm is outlined in Algorithm~\ref{alg:de}.

\begin{algorithm}[!htb]
    \caption{Differential Evolution (DE)}
    \label{alg:de}  
    \algsetup{linenosize=\footnotesize}
        \footnotesize
    	\begin{algorithmic}
    	    \REQUIRE $f : \mathbb{R}^D \to \mathbb{R}$, $CR \in [0, 1]$, $F \in [0, 2]$, $NP \geq 4$
    	    \STATE initialize $NP$ points $\mathcal{P}$ uniformly at random in the search space 
    		\FOR{generation $m=1, 2, \dots$}
    		   \FOR{agent $\mathbf{x} \in \mathcal{P}$}
    		    \STATE randomly select three candidates $\mathbf{a}, \mathbf{b}, \mathbf{c} \in \mathcal{P}$ that are distinct from each other and from $\mathbf{x}$ 
    		    \STATE pick a random index $R \in \lbrace 1, 2 \dots D \rbrace$
    		    \FOR{$i= 1, 2 \dots D$}
    		        \STATE pick $r_i \sim U(0,1)$
    		        \IF{$r_i < CR$ or $i = R$}
    		            \STATE set $y_i = a_i + F (b_i - c_i)$
    		        \ELSE
    		            \STATE set $y_i = x_i$
    		        \ENDIF
    		       \IF{$f(\mathbf{y})\leq f(\mathbf{x})$}
    		        \STATE replace $\mathbf{x}$ in $\mathcal{P}$ with $\mathbf{y}$
    		       \ENDIF
    		    \ENDFOR
    		   \ENDFOR
    		\ENDFOR
    		\STATE return the agent in $\mathcal{P}$ with the least function value
    	\end{algorithmic}
    \end{algorithm}

Here, $CR$ is the crossover probability and denotes the probability of replacing a coordinate in an existing solution with the candidates (crossover), $F$ is the differential weight and controls the strength of the crossover, and $NP$ is the size of the population (larger implies a more global search). We use standard values $CR = 0.7, F = 0.5$ and $NP = 32$ for reporting experimental results, unless indicated otherwise. The search bounds are also set to $[-4, +4]$ for all coordinates; all points that are generated during the initialization of the population and the crossover are clipped to this range. 

\paragraph{Experimental Details:} We consider both transfer and multi-task learning settings. For the transfer experiment, we use the Rosenbrock, Ackley and Sphere functions as the source tasks, and the Rastrigin function as the target task. We also consider each of the source tasks as the ground truth to see whether we can identify the correct source task in an ideal controlled setting. In the multi-task setting, we did not observe any advantages of our algorithm nor the baselines when solving the Rosenbrock, Ackley or Sphere functions, so we focus on the quality of solution obtained for the Rastrigin function only. \textbf{Transfer learning:} We first generate demonstrations $(\mathbf{x}_i, f(\mathbf{x}_i))$ from each source function $f$ using DE (Algorithm~\ref{alg:de}) and configuration above until a fitness of $0.15$ is achieved. Respectively, these have sizes 17693, 6452 and 5853 for each source task. We further transform outputs for training and prediction with the neural-linear model using a log-transform $y \mapsto \log(1 + y)$ to limit the effects of outliers and stabilize the training of the model. When solving the target task, the batch $\mathcal{B}$ is a singleton set containing the best solution from the source data, and $O_{base}$ is trained by replacing the first of the three sampled candidates prior to crossover with probability $p_m = 0.99^m$, where $m$ is the index of the current generation (following the structure of Algorithm~\ref{alg:main}). This guarantees that the new swarm will possess the traits of the source solutions with high probability, but still maintain diversity so the solution can be improved further. \textbf{Multi-task learning:} We solve all source and target functions simultaneously, sharing the best solutions between them using the mechanisms outlined above for the transfer learning experiment. One QP is maintained for each task to learn weightings over the other tasks excluding itself. Here, we set $p_m = 0.3$, so that the best obtained solutions can be consistently shared between tasks over time. 

\subsection*{Details for the Supply Chain Benchmark}

\paragraph{Problem Setting:}
Demand in units per time step is $d_{t,i} \sim \mathrm{Poisson}(\mu_i)$, where $\mu_i$ is $\lbrace 7, 6, 6, 5, 5, 5 \rbrace$ for warehouses (factory demand is zero) and is not backlogged but lost forever during shortages. The unit selling price is $0.6$, the unit production cost is $0.1$, the unit storage cost per time step is $0.03$ and the maximum storage capacity is capped at $50$ for the factory and each warehouse. A truck can ship only $4$ units, but the controller can dispatch an unlimited number of trucks from any location. 

\paragraph{Solver Settings:}
We use the Deep Deterministic Policy Gradient (DDPG) Algorithm \citep{lillicrap2015continuous} to solve this problem. The critic network has 300 hidden units per layer. We also use L2 penalty $\lambda = 10^{-4}$, learning rates $10^{-4}$ and $2 \times 10^{-4}$ for actor and critic respectively, $U(-3\times 10^{-3}, 3\times 10^{-3})$ initialization for weights in output layers, discount factor $\gamma = 0.96$, horizon $T = 200$, randomized replay with capacity $10000$, batch size of $32$, and explore using independent Gaussian noise $\mathcal{N}(0, \sigma_t^2)$, where $\sigma_t$ is annealed from $0.15$ to $0$ linearly over $50000$ training steps. 

\paragraph{Experimental Details:} We consider the transfer from multiple source tasks (Scenarios 1, 2 and 3) to a single target task (Target Task) as indicated in Figure~\ref{fig:source_target_supply}. We also consider each source task as a ground truth. To collect source data, we train DDPG with the above hyper-parameters on each source task for $30000$ time steps, then randomly sub-sample $10000$ observations. This last step demonstrates whether our approach can learn stable representations with limited data and incomplete exploration trajectories. Since regression is sensitive to outliers, when training the neural linear model, we remove $2.5\%$ of the samples with the highest and lowest rewards (we also exclude observations collected from the target task that lie outside any of these bounds). In order to implement transfer, we set $p_m = 0.95^m$, where $m$ is the current episode number. This provides a reasonable balance between exploration of source data and exploitation of target data. In accordance with Algorithm~\ref{alg:main}, we sample batches $\mathcal{B}$ of size 32 uniformly at random from the source data. 

\subsection*{Details for Prioritized Experience Replay}

\paragraph{Experimental Details:}
We evaluated the performance of Prioritized Experience Replay \citep{schaul2015prioritized} (PER) on the Supply Chain benchmark. In summary, PER is a replay buffer that ranks experiences using the Bellman error, defined for the DDPG algorithm for a transition $(s,a,r,s')$ as
\begin{equation*}
    \delta = r + \gamma Q'(s', \mu'(s')) - Q(s, a),
\end{equation*}
where $Q$ is the critic network, and $Q'$ and $\mu'$ are the target critic and target actor networks, respectively.

In order to use PER to transfer experiences, we load all source demonstrations into the replay buffer prior to target task training with initial Bellman error $\delta = 1$. The source demonstrations are first combined into a single data set and shuffled. The capacity of the buffer is then fixed to the total number of source demonstrations from all tasks (around 28,000 examples). During target training, observed transitions are immediately added to the replay buffer and override the oldest experience. In this way, the agent is able to maximize the use of the source data at the beginning of training but eventually shift emphasis to target task data.

\begin{figure}[!htb]
    \centering
    \begin{subfigure}{.495\textwidth}
        \centering
        \includegraphics[width=0.99\textwidth]{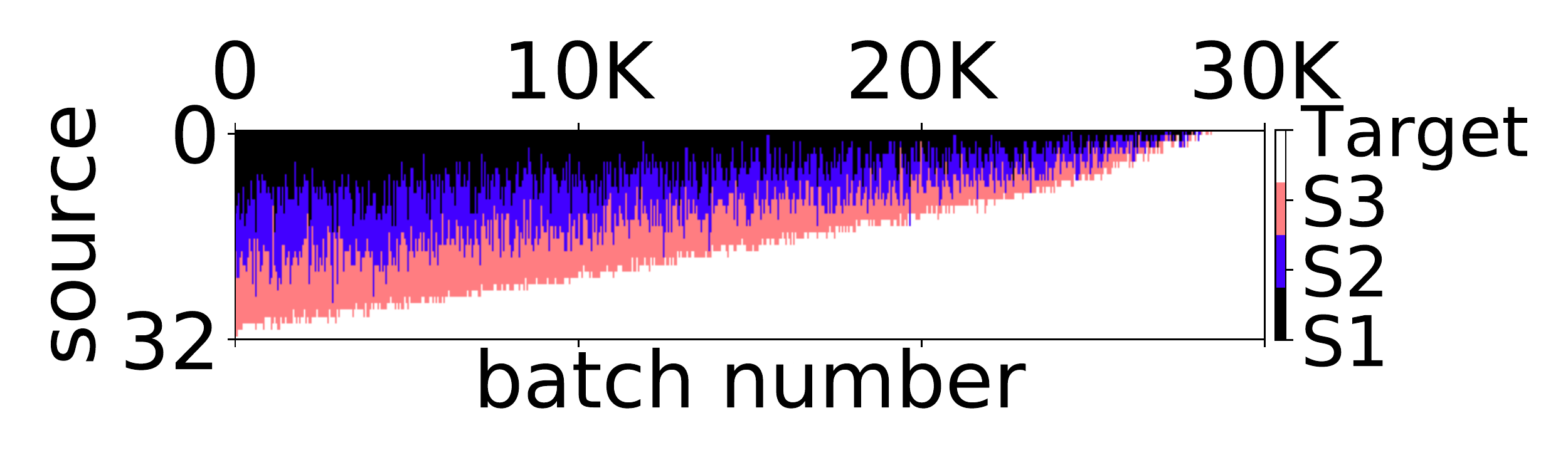}
        \caption{Scenario 1}
    \end{subfigure}%
    \begin{subfigure}{.495\textwidth}
        \centering
        \includegraphics[width=0.99\textwidth]{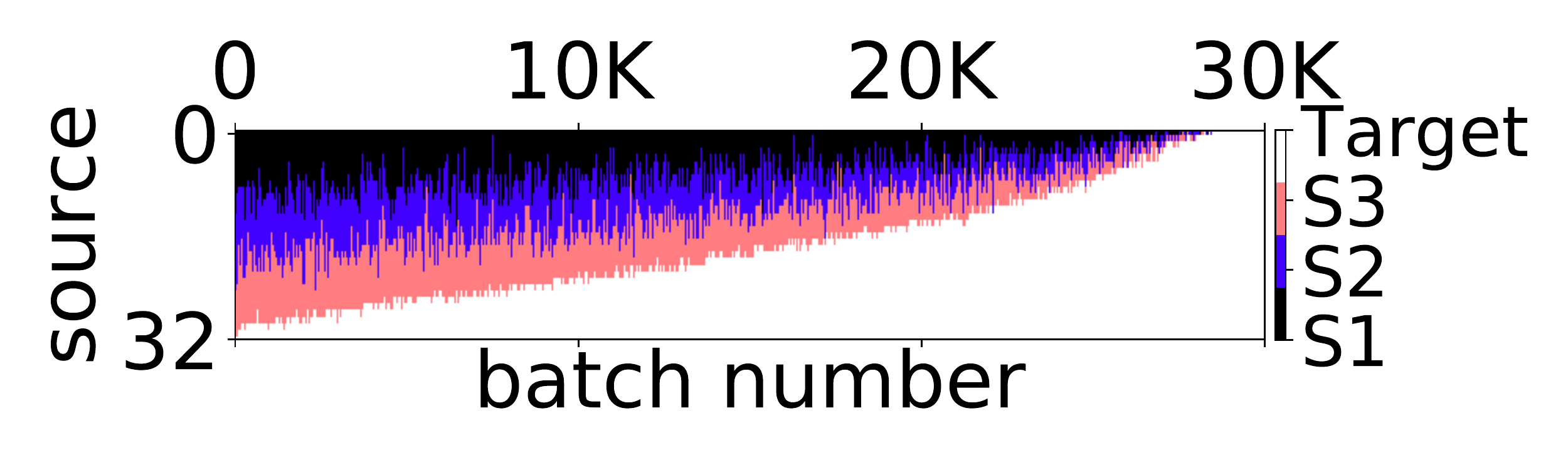}
        \caption{Scenario 2}
    \end{subfigure}
    \begin{subfigure}{.495\textwidth}
        \centering
        \includegraphics[width=0.99\textwidth]{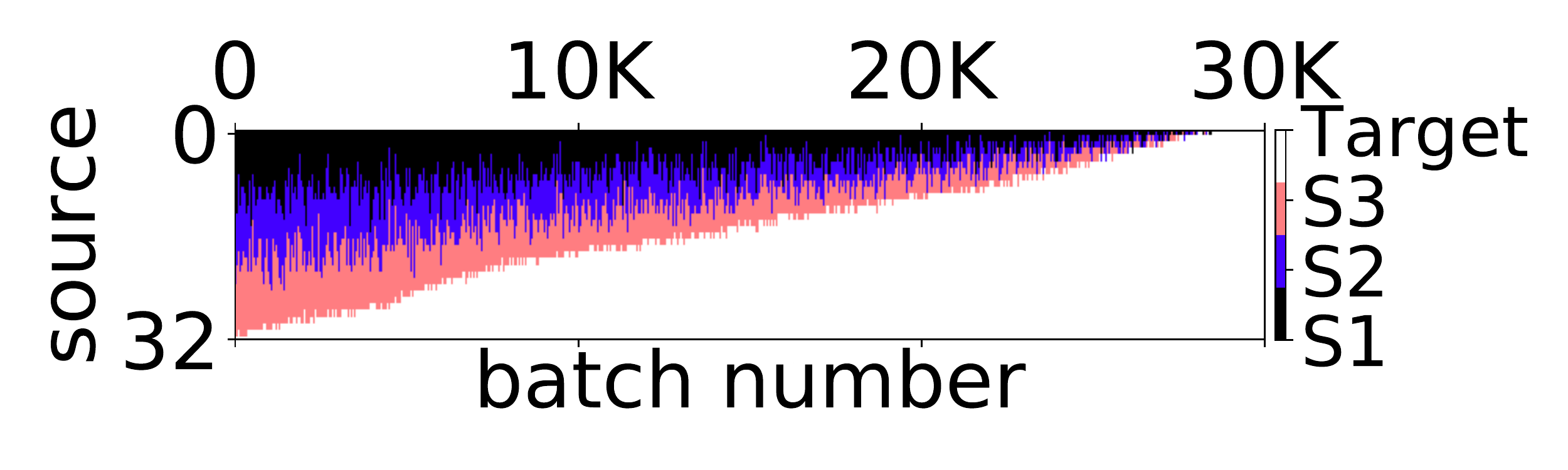}
        \caption{Scenario 3}
    \end{subfigure}%
    \begin{subfigure}{.495\textwidth}
        \centering
        \includegraphics[width=0.99\textwidth]{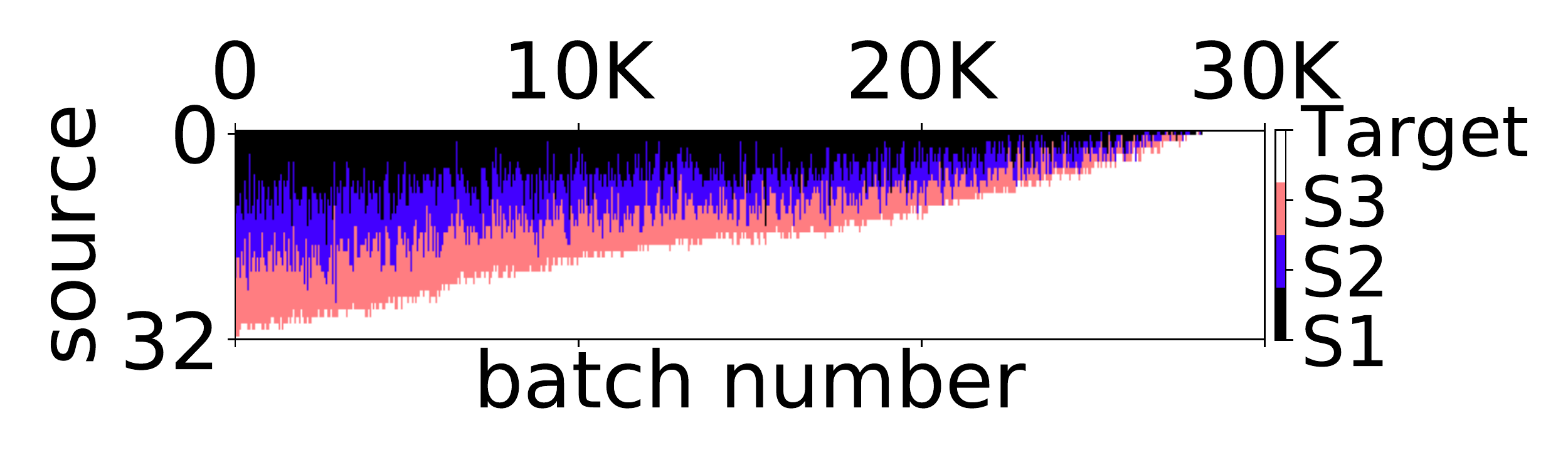}
        \caption{Scenario 4}
    \end{subfigure}
    \caption{Shows the composition of each batch over time sampled from the prioritized replay (PER) according to whether the sample came from the source or target task data, for each ground truth. PER is unable to rank experiences correctly to match the context.}
    \label{fig:per_structure}
\end{figure}

\paragraph{Detailed Analysis:}
Figure~\ref{fig:per_structure} demonstrates the composition of each batch (of size 32) according to source (Scenario 1, Scenario 2, Scenario 3, Target Task) for each ground truth. As illustrated in all four plots, in early stages of training, batches are predominantly composed of source data, while in later stages, they consist entirely of target data. Figure~\ref{fig:per_structure} shows that PER is unable to favor demonstrations from the source task that correspond to the ground truth, in all four problem settings. One possible explanation of this is that an implicit assumption of PER is violated, namely that experiences are drawn from the same distribution of rewards, whereas in our experiment, experiences can come from tasks with contradictory goals. In this case, experiences corresponding to the ground truth do not necessarily have larger Bellman error (in fact, the opposite may be true).  

\end{document}